\documentclass[journal]{IEEEtran}
\usepackage{booktabs}
\usepackage{tabularx}
\usepackage{amsmath,amssymb,amsfonts}
\usepackage{algpseudocode}
\usepackage{algorithm}
\usepackage{array}
\usepackage{uniquecounter}
\usepackage[final]{hyperref}
\usepackage{textcomp}
\usepackage{stfloats}
\usepackage{url}
\usepackage{verbatim}
\usepackage{graphicx}
\usepackage{cite}
\usepackage{cleveref}
\usepackage{xcolor}
\usepackage{subcaption}
\usepackage{diagbox}
\usepackage{multirow}

\newcommand{\result}[2]{#1 - #2}

\begin{document}

\title{Adaptive Semantic Token Communication \\for Transformer-based Edge Inference \vspace{.3cm}}

\author{ Alessio Devoto$^*$, Jary Pomponi$^*$, Mattia Merluzzi,\\ Paolo Di Lorenzo, Simone Scardapane 
%
\thanks{A. Devoto is with the Department of Computer, Control, and Management Engineering (DIAG) at Sapienza University of Rome, Italy. He is also a member of the Consorzio Nazionale Interuniversitario per le Telecomunicazioni (CNIT), Parma, Italy. Jary Pomponi, Paolo Di Lorenzo, and Simone Scardapane are with the Department of Information Engineering, Electronics, and Telecommunications (DIET) at Sapienza University of Rome, Italy, and they are also affiliated with CNIT. Mattia Merluzzi is with CEA-Leti, Université Grenoble Alpes, located in Grenoble, France. Email:\{alessio.devoto, jary.pomponi, paolo.dilorenzo, simone.scardapane\}@uniroma1.it, mattia.merluzzi@cea.fr. 

This work has been supported by the SNS JU project 6G-GOALS under the EU’s Horizon program Grant Agreement No 101139232, by Sapienza grant RG123188B3EF6A80 (CENTS), and by European Union under the Italian National Recovery and Resilience Plan of NextGenerationEU, partnership on Telecommunications of the Future (PE00000001 - program RESTART). We also acknowledge ISCRA for awarding this project access to the LEONARDO supercomputer, owned by the EuroHPC Joint Undertaking, hosted by CINECA (Italy).}\vspace{-.4cm}
}

\markboth{Journal of \LaTeX\ Class Files,~Vol.~14, No.~8, August~2021}%
{Shell \MakeLowercase{\textit{et al.}}: A Sample Article Using IEEEtran.cls for IEEE Journals}


\maketitle
\begingroup\renewcommand\thefootnote{*}
\footnotetext{Equal contribution}
\endgroup

\vspace{-.8cm}

\begin{abstract}
This paper presents an adaptive framework for edge inference based on a dynamically configurable transformer-powered deep joint source channel coding (DJSCC) architecture. Motivated by a practical scenario where a resource constrained edge device engages in goal oriented semantic communication, such as selectively transmitting essential features for object detection to an edge server, our approach enables efficient task aware data transmission under varying bandwidth and channel conditions. To achieve this, input data is tokenized into compact high level semantic representations, refined by a transformer, and transmitted over noisy wireless channels. As part of the DJSCC pipeline, we employ a semantic token selection mechanism that adaptively compresses informative features into a user specified number of tokens per sample. These tokens are then further compressed through the JSCC module, enabling a flexible token communication strategy that adjusts both the number of transmitted tokens and their embedding dimensions. We incorporate a resource allocation algorithm based on Lyapunov stochastic optimization to enhance robustness under dynamic network conditions, effectively balancing compression efficiency and task performance. Experimental results demonstrate that our system consistently outperforms existing baselines, highlighting its potential as a strong foundation for AI native semantic communication in edge intelligence applications.

\end{abstract}

\section{Introduction}
\IEEEPARstart{R}{ecently}, there was a surge of interest in semantic and goal-oriented communications, establishing this paradigm as crucial for the development of 6G AI-native communication networks
\cite{strinati20216g, strinati2024goal}. 
In contrast to traditional communication systems, which focus on efficiently transmitting the data in the form of raw bits, semantic communication emphasizes understanding the meaning and intent behind the data, allowing for more efficient information exchange tailored to the specific goals of the communication task. This prioritizes the meaning and purpose behind transmitted data instead of the transmission itself and reduces the need for exact bit-level reconstructions, thereby offering gains in efficiency and adaptability, particularly in dynamic or resource constrained environments \cite{10183788, qiao2025token, binucci2024enabling, 2025arXiv250312484C,di2023goal}.
%

A prominent use case for semantic communication is edge inference, where distributed edge devices collect high dimensional sensor data and must convey only the most task relevant information to a nearby server for real time decision making \cite{di2023goal}. In such scenarios, goal-oriented semantic communication enables these devices to transmit compressed, high level representations that are sufficient for the target inference task, such as object detection or anomaly recognition, while significantly reducing communication overhead. This is especially important in wireless edge networks, where constraints on bandwidth, latency, and power demand intelligent mechanisms to select and transmit only semantically meaningful features aligned with the task objective. In this context, transformer-based neural networks, with their inherent attention mechanisms and contextual processing capabilities, serve as ideal candidates for extracting and compressing semantic information from data into a lower-dimensional representation space \cite{xie2021deep}. This representation consists of multiple vectors called \textit{tokens}, each capturing different semantic aspects of the input. By treating tokens as discrete transmission units, the system can selectively transmit only the most semantically relevant tokens while discarding less informative ones; this property has enabled the development of semantic token-based communication systems that adapt to dynamic channel conditions and application requirements, see, e.g. \cite{devoto2024adaptive,qiao2025token,qiao2025token2}.


\subsection{Related works} 
Edge inference comes with the drawback of an ad-hoc deployment of computing resources, which might be more costly in terms of economic and environmental aspects, if compared to a shared infrastructure. However, it allows end devices and service consumers to keep data local, thus promoting privacy and secrecy, and retrieving results within strict latency constraints, without the need of reaching distant central clouds. To achieve a positive balance between benefits and drawbacks, the joint optimization of communication and computing segments is a fundamental brick~\cite{di2023goal}. In general, edge inference can take place: $i)$ locally at an end device, $ii)$ fully offloaded to an edge server (or, Mobile Edge Host-MEH), or $iii)$ partially performed at the two end sides~\cite{Matsubara22}. The first solution (full local inference) typically requires model optimization techniques such as pruning or quantization \cite{Wangpruning2024}, to be able to efficiently run a full (deep) model locally in a resource limited device. This comes with no communication overhead and full secrecy as no data is exposed, however at a cost of local computation and possibly reduced inference accuracy. The second solution (full offloading) \cite{Zichuan21, Quan24} relieves the device from any processing, but it can lead to high communication overhead and data exposure.  In the third case (partial offloading), which is the most flexible one, as also highlighted in \cite{Lee23, binucci2024enabling}, part of the computation is performed locally to extract relevant (possibly lower dimensional) features. This is helpful if the edge server is temporary unavailable, or to reduce the communication overhead needed to offload the inference task. 

Several methodologies have been proposed to enable local feature extraction with the goal of minimizing communication overhead while maintaining acceptable task performance, typically measured in terms of accuracy. For a comprehensive overview, the reader is referred to the recent survey in~\cite{ZhangSEC24}. One straightforward approach is to split a deep neural network (DNN) between a local device and an edge server, a technique commonly referred to as DNN splitting~\cite{Matsubara22}. This splitting can be adapted dynamically based on channel conditions and the availability of radio and computational resources~\cite{Lee23, Labriji23}. From the perspective of task performance (e.g., achieving a target classification accuracy), prior work has shown that features extracted at different layers—corresponding to different splitting points—exhibit varying levels of robustness to channel noise and transmission errors~\cite{binucci2024enabling}. This observation highlights the importance of selecting optimal split points depending on system constraints. When scaling to powerful architectures such as Vision Transformers, the opportunity for DNN splitting is further enriched by mechanisms like token pruning, which can significantly reduce communication costs by discarding less informative tokens. Moreover, when the model is trained end-to-end with the wireless channel modeled as a differentiable layer, DNN splitting effectively performs joint source and channel coding. This results in a learned \textit{Deep Joint Source-Channel Coding} (DJSCC) scheme that simultaneously achieves compression and error resilience, optimized for the specific communication environment.

Recent advancements in wireless semantic communication have highlighted DJSCC as a promising approach, see, e.g., \cite{xu2023deep, gunduz2024joint}. DJSCC systems integrate source and channel coding into a unified process over wireless channels, employing deep neural networks to tailor transmission strategies to specific application requirements. By modeling the communication channel as a differentiable layer within the network, these systems enable end-to-end learning that jointly adapts both source and channel coding, leading to improved performance across key metrics. DJSCC methods have shown clear advantages over traditional coding schemes, particularly in semantic communication settings, by exhibiting graceful performance degradation below information-theoretic signal-to-noise ratio (SNR) thresholds and avoiding the sharp declines characteristic of conventional approaches \cite{bourtsoulatze2019deep,lo2023collaborative}. This makes DJSCC a compelling choice for practical scenarios characterized by low bandwidth and/or low SNR, where traditional methods struggle to maintain reliable performance.

Most existing JSCC systems operate under fixed trade-offs 
optimized for specific channel conditions and may struggle to generalize across different or fluctuating environments. However, in dynamic communication settings, where channel conditions can vary rapidly, the need for adaptive mechanisms that can adjust their operating parameters in real time becomes essential. To this end, recent studies have introduced adaptive methods that dynamically respond to real-time channel variations. These approaches include partitioning the model's latent space to support variable transmission rates \cite{xu2023deep}, masking non-essential channels within convolutional architectures \cite{yang2022deep}, and employing transformer-based architectures to project tokens onto variable-length representations \cite{dai2022nonlinear}. Similarly, the model in \cite{yang2024swinjscc} demonstrates superior coding efficiency compared to CNN-based JSCC and traditional BPG + LDPC schemes, while achieving lower end-to-end latency. Furthermore, \cite{bian2025process} leverages transformer-based models in cooperative relay networks, showcasing notable beamforming gains. 
Additionally, several works have focused specifically on communication efficiency. For example, Wang et al. \cite{wang2023adaptive} developed an adaptive source allocation strategy based on quantization, enhancing transmission reliability under changing conditions. Similarly, Zhou et al. \cite{zhou2021semantic} introduced a Universal Transformer for semantic text transmission that adjusts its communication protocol depending on the channel environment. Finally, the work in \cite{binucci2023multi} proposed a semantic-aware edge learning framework for multi-user wireless networks, where users adaptively decide between local and remote inference and transmit only task-relevant features via goal-oriented compression, while a Lyapunov-based optimization dynamically allocates communication and computational resources to balance energy, latency, and inference accuracy.

An additional degree of freedom in the design of DJSCC involves rendering the model adaptive, enabling it to dynamically adjust its computational complexity and output dimensionality, thereby enhancing flexibility and performance across diverse operating conditions \cite{xu2023deep}. Adaptive computation can be achieved through several approaches, broadly categorized into two strategies: i) employing multiple specialized models for different trade-offs \cite{courbariaux2014training, wu2020integer, dettmers2022llm, aguilar2020knowledge, hinton2015distilling, hoefler2021sparsity}; or ii) embedding flexibility into a single model capable of handling a range of constraints. Here, we focus on the second scenario. Models in this category embed adaptivity into model architectures via dynamic networks that adjust processing based on input complexity \cite{condcomp_tutorial}. Some methods employ sampling to process only relevant image regions within budget constraints \cite{avit, han2021dynamic, devoto2024alast}, while others activate specific submodules conditionally \cite{wojcik2023adaptive}. Mixture-of-Experts approaches dynamically route inputs through specialized sub-networks \cite{foedus2021, shazeermoe, moefication}, optimizing resource use. Early Exit techniques introduce intermediate classifiers to halt computation once sufficient confidence is achieved \cite{pomp2024, zhou2020bert, han2021dynamic, krzepkowski2024joint}.

\subsection{Contributions of the Paper} 

In this work, we propose a novel transformer based DJSCC algorithm tailored for edge inference scenarios, where resource constrained devices must transmit task relevant information under dynamic channel conditions and limited computational budgets. Building on the conditional computation framework introduced in \cite{condcomp_tutorial} and extending our preliminary work in \cite{devoto2024adaptive}, we introduce a trainable \textit{semantic token selection} mechanism that leverages the structural properties of transformer models to dynamically identify and transmit the most relevant input components, such as image patches in vision tasks, based on relevance to the downstream task.

The proposed architecture offers three key advancements: (i) it compresses semantic tokens into a complex valued representation space while minimizing information loss during reconstruction at the receiver; (ii) it achieves robustness across a wide range of channel noise conditions using a single model, removing the need for multiple specialized DJSCC networks; and (iii) it operates in the complex domain while remaining compatible with standard real valued neural networks, allowing seamless integration with pre trained models. By reducing the number of tokens processed and transmitted, the semantic token selection mechanism also reduces the overall complexity of the DJSCC model, improving its efficiency and scalability, particularly in resource constrained environments. Additionally, we incorporate a dynamic resource allocation algorithm based on stochastic optimization \cite{neely2010stochastic}, which adaptively tunes system parameters to maximize inference performance while accounting for communication overhead in time varying conditions. Integrated with our DJSCC framework, this mechanism enables real time adaptation of token selection and compression, improving the semantic efficiency of transmitted information in support of edge inference tasks. Through comparative analysis with existing DJSCC methods and conventional communication schemes, we demonstrate that our system achieves superior accuracy compression trade offs in edge inference settings by leveraging its additional degrees of freedom. The proposed token selection mechanism also facilitates the extraction of semantically rich and interpretable representations, laying the groundwork for interpretable by design models in next generation AI native communication systems. Its adaptability to varying channel conditions and task requirements makes it particularly well suited for intelligent edge devices, where efficient resource management is critical to achieving high performance inference.

\noindent\textbf{Outline.} This paper is structured as follows: Section II provides background on Vision Transformers (ViT), highlighting their token-based architecture and suitability for semantic communication. Section III introduces the system model, defining the communication framework between edge devices and servers. Section IV details the design of the proposed token-based DJSCC architecture, including adaptive token selection and compression mechanisms. Section V presents a dynamic optimization strategy based on Lyapunov stochastic control for real-time adaptation to varying channel conditions under bandwidth and noise constraints. Section VI reports empirical results that validate the performance and adaptability of the proposed system under different SNR regimes, offering qualitative insights into the semantic interpretability of token selection. Finally, Section VIII draws the conclusions and directions for future work.

\section{Background on Vision Transformers}
\label{sec:background}
%
In this section, we provide a brief overview of the Vision Transformer (ViT) model \cite{dosovitskiy2021an}, which serves as a core component of our proposed token-based communication framework, although the method can be readily adapted to scenarios involving other transformer-based architectures. We summarize recurring notation in Table \ref{tab:symbols}.

\begin{table}[t]
\centering
\scriptsize
\caption{List of recurring symbols}
\begin{tabularx}{0.65\linewidth}{>{\raggedright\arraybackslash}p{1.8cm}X}
\toprule
\textbf{Symbol} & \textbf{Description} \\
\midrule
$\mathbf{x}$ & Input image of size $\mathbb{R}^p$ with $p= {C\cdot H \cdot W}$ \\
$ \mathcal{E}, \mathcal{B}, \mathcal{C}$ & ViT blocks (resp. preprocessing, transformer block, classification head) \\
$n$ & Number of initial tokens \\
$\alpha$, $n_\alpha$ & User-defined budget, number of selected tokens \\
$d$ & Original dimension of each token \\
$o$, $r$ & Compressed dimension $o < d$, with ratio $r=\frac{o}{d}$ \\
$\mathbf{H}^l$ & Set of tokens at layer $l$ of the ViT, of shape $n \cdot d$ \\
$\mathbf{s}$ & Symbols transmitted on the channel \\
$E(\mathbf{x})$, $D(\mathbf{s})$ & DJSCC encoder and decoder \\
$\eta$ & Channel (non-trainable) \\
$\mathcal{C}_E$, $\mathcal{C}_D$ & Receiver encoder and decoder \\
$\boldsymbol{\gamma}(t)$ & Optimized time-varying parameters $\{r(t), \alpha(t)\}$ \\
\bottomrule
\end{tabularx}
\label{tab:symbols}
\end{table}

A ViT model, denoted as $f(\mathbf{x})$, takes an image $\mathbf{x} \in \mathbb{R}^{C\cdot H \cdot W}$ as input, where $C$, $H$, and $W$ represent the number of channels, height, and width of the image, respectively. The overall structure of the model is defined as follows:
\begin{equation}\label{eq:VIT}
    f(\mathbf{x}) = \mathcal{C} \circ \mathcal{B}^L \circ \mathcal{B}^{L-1} \circ \cdots \circ \mathcal{B}^{1} \circ \mathcal{E}(\mathbf{x}).
\end{equation}
In \eqref{eq:VIT}, $\mathcal{E}(\mathbf{x})$ is a preprocessing layer that turns the image into a sequence of tokens $\mathbf{H}^0 \in \mathbb{R}^{n \times d}$, with $n$ the number of tokens and $d$ their features; then, $\{\mathcal{B}^1 \dots \mathcal{B}^{L}\}$ are transformer blocks that process the tokens via multi-head attention (MHA) and feed-forward networks, with $L$ denoting the number of blocks. The set of tokens is created by dividing an image into non-overlapping patches of fixed length, which are then flattened and projected to a fixed embedding size (i.e., $d$) using a trainable network.  To preserve spatial information, a unique learnable-positional embedding is added to each token, encoding the original position of each patch within the image. Finally, the tokens set includes a trainable class token, which is used as input to the classification layer $\mathcal{C}$ to produce the final prediction vector. 

The MHA mechanism represents the core of each transformer block. This mechanism consists of $H$ parallel self-attention heads, each computing outputs by analyzing interactions between tokens in the input sequence. These outputs are then concatenated along the feature dimension and projected through a learnable matrix $\mathbf{W}_o \in \mathbb{R}^{Hd_v \times d}$, where $d_v$ is the output dimension of each attention head and $d$ is the model’s hidden dimension. Letting $\mathbf{H}^{l-1} \in \mathbb{R}^{n \times d}$ denote the sequence of token embeddings input to the $l$-th transformer block, the MHA function reads as:
\begin{equation*}\label{eq:MHA}
  \text{MHA}^l(\mathbf{H}^{l-1}) = \left[\text{SA}_1(\mathbf{H}^{l-1}), \dots, \text{SA}_H(\mathbf{H}^{l-1})\right] \mathbf{W}_o  
\end{equation*}
for $l=1,\ldots,L$, where $\text{SA}_i$ is the $i$-th Self-Attention head:
\begin{equation*}\label{eq:SA}
\text{SA}_i(\mathbf{H}^{l-1}) = \text{softmax}\left(\frac{\mathbf{Q}_i \mathbf{K}_i^\top}{\sqrt{d_k}}\right)\mathbf{V}_i,
\end{equation*}
for $i=1,\ldots,H$, where the softmax is applied row-wise. Each attention head in (\ref{eq:SA}) operates using three learned projections: Query ($\mathbf{Q}_i$), Key ($\mathbf{K}_i$), and Value ($\mathbf{V}_i$), computed as follows:
\begin{align*}
\mathbf{Q}_i &= \mathbf{H}^{l-1}\mathbf{W}_{q,i}, & \mathbf{W}_{q,i} &\in \mathbb{R}^{d \times d_k},\\
\mathbf{K}_i &= \mathbf{H}^{l-1}\mathbf{W}_{k,i}, & \mathbf{W}_{k,i} &\in \mathbb{R}^{d \times d_k},\\
\mathbf{V}_i &= \mathbf{H}^{l-1}\mathbf{W}_{v,i}, & \mathbf{W}_{v,i} &\in \mathbb{R}^{d \times d_v},
\end{align*}
where $d_k$ is the dimension of the queries and keys. The resulting matrices $\mathbf{Q}_i$, $\mathbf{K}_i$, and $\mathbf{V}_i$ have dimensions $(n, d_k)$, $(n, d_k)$, and $(n, d_v)$ respectively. After computing all $H$ attention outputs, they are concatenated to form a matrix of shape $(n, Hd_v)$ in (\ref{eq:MHA}), which is then linearly projected back to dimension $d$ using $\mathbf{W}_o$. A key observation is that while each self-attention output preserves the dimensions of its input, only the feature dimension $d$ must remain fixed throughout the model. The token count $n$, however, can be dynamically adjusted without compromising the MHA mechanism.

In the sequel, we exploit this flexibility to develop a token selection mechanism that identifies and preserves only the most semantically relevant tokens based on a user-defined computational budget and input characteristics. This approach offers three significant advantages. First, the computational burden is reduced, as the model processes a smaller number of tokens. Second, the selected tokens can be interpreted as the most meaningful part of the input image, thus providing useful insights into the model’s internal reasoning. Last, in a semantic communication setting, transmitting only the most informative tokens enhances communication efficiency by reducing bandwidth requirements while preserving the core semantic content, enabling more reliable and meaningful exchanges between sender and receiver. 

\label{sec:sys_model}
\begin{figure}[t]
    \centering
    \includegraphics[width=.9\linewidth]{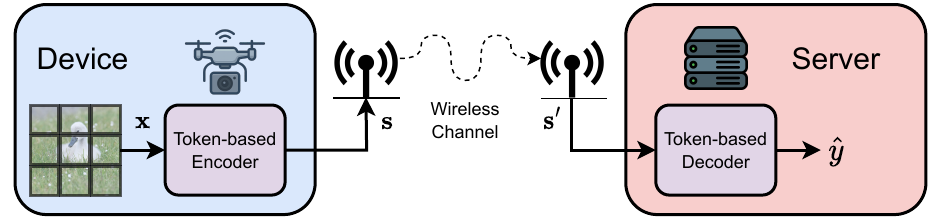}
    \caption{System scenario for semantic-oriented edge inference.}
    \label{fig:problem}
\end{figure}

\section{System Model}

We consider a communication scenario with a single edge device equipped with a camera that captures images represented by arrays \( \mathbf{x} \in \mathbb{R}^p \), with $p=C\cdot H\cdot W$. The device applies a token-based DJSCC function \( E: \mathbb{R}^p \rightarrow \mathbb{C}^q \) to produce the transmitted signal \( \mathbf{s} = E(\mathbf{x}) \in \mathbb{C}^q \), subject to an average power constraint:
\begin{equation*}\label{eq:Power_constraint}
 \frac{1}{q} \| \mathbf{s} \|_2^2 \leq 1,   
\end{equation*}
where \( q \) denotes the available number of transmitted symbols. The compression ratio of the system is defined as: 
\begin{equation*}\label{eq:compression_ratio}
\rho = \frac{q}{p},    
\end{equation*}
capturing the fraction of the input dimensionality that is retained in the transmitted representation. This allows the encoder to adapt the level of compression to meet resource constraints while preserving task-relevant semantics. Then, the signal \( \mathbf{s} \) is transmitted over a wireless channel to a nearby edge server, which performs a downstream inference task such as classification or retrieval. At the server side, a token-based deep joint source channel decoder \( D: \mathbb{C}^q \rightarrow \mathcal{Y} \) maps the received signal to an output \( \hat{y} \in \mathcal{Y} \), where \( \mathcal{Y} \) denotes the output space (e.g., class labels). \Cref{fig:problem} provides an illustration of the proposed edge communication setting.

\subsubsection*{Channel Model}
Communication occurs over a point-to-point complex additive white Gaussian noise (AWGN) channel. The received signal is given by
\begin{equation*}\label{eq:channel}
\mathbf{s}' = h \mathbf{s} + \mathbf{n},    
\end{equation*}
where \( h \in \mathbb{C} \) is the channel gain and \( \mathbf{n} \in \mathbb{C}^q \) is an additive noise vector with entries drawn i.i.d. from a complex normal distribution \( \mathcal{CN}(0, \sigma_n^2) \). For the AWGN channel, we set \( h = 1 \). Under a slow fading model, \( h \) is sampled independently from \( \mathcal{CN}(0, \sigma_h^2) \) for each inference task and remains constant during a single task. The signal-to-noise ratio (SNR) for the  communication of vector $\mathbf{s}$ can then be expressed as ${\rm SNR}=|h|^2 \|s\|^2/\sigma_n^2$.

\section{Token-based DJSCC Design}
\label{sec:method}
%

\begin{figure*}[t!]
\centering
  \includegraphics[width=\textwidth]{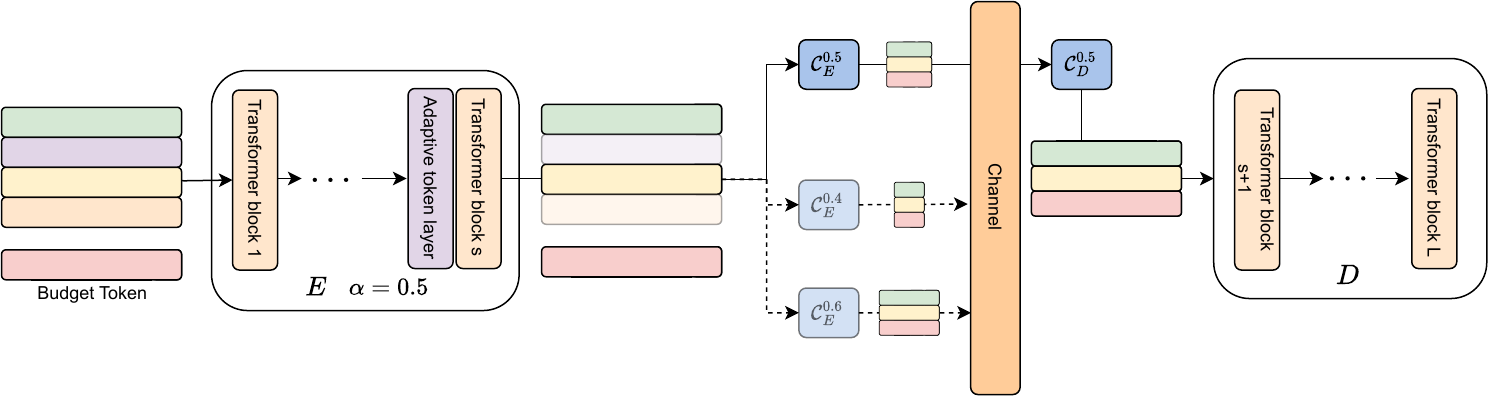}
  \caption{Schema of the proposed token-based DJSCC. 
  } \label{fig:general-pipeline}
\end{figure*}

In this section, we introduce our token-based DJSCC architecture, which incorporates an adaptive token selection and compression mechanism. This design enables dynamic control over output resolution, computational cost, and communication load by selectively retaining and transmitting only the most informative tokens. Specifically, we consider a DJSCC scenario where the entire communication pipeline is described by an end to end deep neural network composed of multiple macro-blocks:
\begin{equation} \label{eq:f_jscc}
f^{\text{jscc}}(\mathbf{x}) = D\circ \eta \circ E(\mathbf{x}),
\end{equation}
where again $\mathbf{x} \in \mathbb{R}^{C\cdot H\cdot W}$ is the input image, while $E$ and $D$ are, respectively, the transmission and the receiver pipelines, and $\eta$ the channel in (\ref{eq:channel}), modeled as a non trainable layer.
Both the transmitter and receiver are constructed based on a pre-trained ViT. Given a ViT with $L$ transformer blocks, as defined in (\ref{eq:VIT}), we partition the model by selecting a splitting point $1 < s < L$. The first $s$ blocks are assigned to the encoder $E$, while the remaining $L - s$ blocks form the decoder $D$. Then, the encoder $E$ is augmented by two novel components: (i) a token selection module, denoted $\mathcal{S}$, is added before each transformer block to identify a subset of semantically relevant tokens; and (ii) a token compression mechanism, denoted $\mathcal{C}_E$, is added before at the end of the encoder to adaptively compress the selected tokens’ features and converts them into complex symbols for transmission over the channel. Mathematically, we have:
\begin{equation}\label{eq:encoder}
E = {\color{red}\mathcal{C}_E} \circ \left(  {\color{red}\mathcal{S}^s}\circ\mathcal{B}^{s} \circ \cdots \circ \mathcal{B}^{2} \circ {\color{red}\mathcal{S}^1} \circ \mathcal{B}^{1} \right) \,\circ\, \mathcal{E}.    
\end{equation}
where we show in red the added components. At the receiver side, the representation is first processed by a receiver decoder $\mathcal{C}_D$, which transforms the received complex symbols back into real-valued features, and is then passed through the remaining transformer blocks, i.e., 
\begin{equation}\label{eq:decoder}
D = \mathcal{C} \circ \mathcal{B}^L \circ \cdots \circ \mathcal{B}^{s+1} \circ {\color{red}\mathcal{C}_D}.     
\end{equation}
where again we highlight in red the components which are added to the pre-trained ViT model. An overview of the proposed token-based DJSCC system is depicted in Fig.~\ref{fig:general-pipeline}. In the following subsections, we describe each component of the proposed pipeline in detail.

\subsection{Adaptive Semantic Token Selection}

We begin our discussion with the token selection module \( \mathcal{S} \) in (\ref{eq:encoder}), which extends the ViT formulation in (\ref{eq:VIT}) by dynamically retaining a subset of tokens at each transformer block \( l=1,\ldots,s \). This selection is governed by a computational budget constraint, specified by a scalar \( \alpha \in (0,1] \), which denotes the proportion of tokens to be retained. In other words, \( \alpha \) controls the percentage of total tokens the model preserves for a given input sample \( \mathbf{x} \). Furthermore, we make explicit the dependence on the trainable parameter vector \( \mathbf{w} \) and the budget constraint $\alpha$ in equation~(\ref{eq:f_jscc}) by writing the function as \( f^{\text{jscc}}_{\mathbf{w}}(\mathbf{x},\alpha)\). Given labeled data pairs $(\mathbf{x}, y)$, the training problem can then be formulated as:
\begin{equation}
\begin{split}
    \label{eq:formulation}
    &\min_{\mathbf{w}}  \quad \mathbb{E}_{(\mathbf{x}, y)}\,\mathcal{L}(f^{\text{jscc}}_{\mathbf{w}}(\mathbf{x},\alpha), y) \quad 
    \\ 
    &\;\;\;\;\text{s.t.}  \;\; \rvert \mathcal{N}_{\mathbf{w}}(\mathbf{x},\alpha) - \alpha \cdot n \rvert \le \epsilon, \;\; \forall \, \alpha \in (0,1],
\end{split}
\end{equation} 
where \( n \) denotes the total number of tokens; \( \mathcal{N}_{\mathbf{w}}(\mathbf{x}, \alpha) \) represents the number of output tokens produced by the model $E$ for a given set of weights \( \mathbf{w} \), budget \( \alpha \), and input sample $\mathbf{x}$; and finally, \( \mathcal{L}(\hat{y}, y) \) represents the classification loss, e.g., cross-entropy. From now on, we omit $\mathbf{w}$ for the sake of clarity. Intuitively, the constraint in (\ref{eq:formulation}) ensures that the number of output tokens remains close to the target proportion $\alpha \cdot n$, within a small error $\epsilon$.

\begin{figure*}[t]
    \centering
    \begin{subfigure}[b]{0.4\textwidth}
        \includegraphics[width=\textwidth]{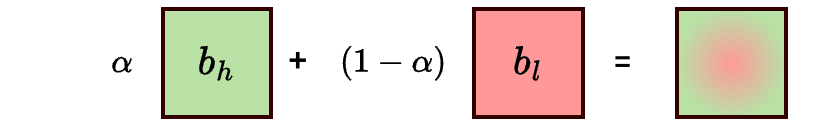}
         \caption{The budget computed as the combination of two trainable tokens corresponding to low ($b_l$) and high ($b_h$) budget. The value $\alpha$ is defined by the user.}
         \label{fig:budget_token}
    \end{subfigure}
    \hfill
    \centering
    \begin{subfigure}[b]{0.55\textwidth}
        \includegraphics[width=\textwidth]{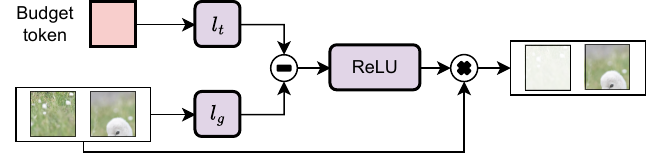}
         \caption{The proposed selection mechanism is applied before each transformer block. The budget token is used to calculate the threshold value, which is evaluated against each gate value to determine if a token must be discarded.}
         \label{fig:zoom_gate}
    \end{subfigure}
    \caption{The details of two components of our proposed token discarding approach. On the left, the creation of the budget token, while on the right, how the tokens are actively discarded. More detail in Section \ref{sec:method}.}
    \label{fig:budget_detail}
\end{figure*}

To solve the constrained problem (\ref{eq:formulation}), we need to design a trainable token selection method that can be incorporated into the ViT model in (\ref{eq:VIT}). To this aim, we introduce a new trainable \textit{budget token} that is appended to the input sequence of image tokens. Specifically, the budget token is formed by combining two trainable tokens: $\mathbf{b}_l \in\mathbb{R}^d$, representing a low-budget configuration, and $\mathbf{b}_h\in\mathbb{R}^d$, representing a high-budget configuration. Then, given a target $\alpha$ value, the budget token is a row-vector computed as:
\[
    \mathbf{b}^0 = \alpha \mathbf{b}_h + (1 - \alpha) \mathbf{b}_l.
\]
When $\alpha \approx 0$, the budget token $\mathbf{b}^0$ approximates $\mathbf{b}_l$, encouraging the model to operate under the lowest possible budget. Conversely, when $\alpha = 1$, the model adopts the high-budget configuration $\mathbf{b}_h$. This behavior is illustrated in Fig.\ref{fig:budget_token}. This additional token is introduced by the preprocessing layer in (\ref{eq:VIT}) as follows:
\[
\mathcal{E}(\mathbf{x}) =
\begin{bmatrix}
\mathbf{H}^0 \\
\mathbf{b}^0
\end{bmatrix}
\in \mathbb{R}^{(n + 1) \times d}
\]
where $\mathbf{H}^0$ represents the initial sequence of tokens. The budget token $\mathbf{b}^0$ is appended to the end of this sequence and propagated through all blocks alongside the other tokens.

\begin{figure}[t]
    \centering
    \includegraphics[width=\linewidth]{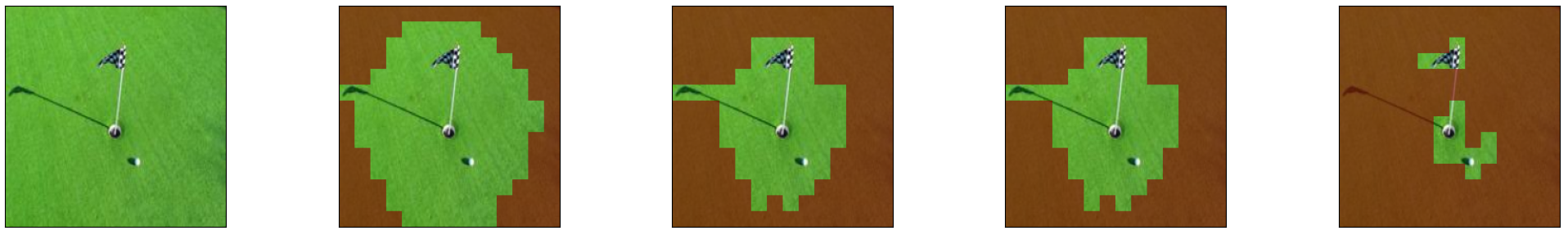}
    \caption{As the images flow through the model (from left to right), some tokens, highlighted in red, are discarded.}
    \label{fig:example_mask}
\end{figure}

For each transformer block, we now introduce a selection mechanism that uses the budget token representation $\mathbf{b}^l$ at the $l$-th layer, along with the current set of tokens, to identify a subset of tokens to discard. A transformer block equipped with this token selection mechanism is referred to as \textit{adaptive}. To this end, we introduce two trainable linear models followed by a sigmoid activation  function, which work together to selectively mask out less important tokens: a threshold model $l_t$ and a token gating model $l_g$. The threshold model $l_t(\cdot) \in [0, 1]$ takes the current budget token $\mathbf{b}^l$ as input and outputs a scalar threshold. The token gating model $l_g(\cdot) \in [0, 1]^{(n - 1)}$ produces a gating score for each token embedding $\mathbf{h}_j^l$, where $\mathbf{h}_j^l$ denotes the $j$-th row of $\mathbf{H}^l$ for $j = 1, \ldots, n - 1$, excluding the budget and class tokens, which are never discarded. These outputs are used to compute a differentiable mask 
 over the input tokens. For each token embedding $\mathbf{h}_j^l \in \mathbf{H}^l$, the corresponding mask value $M_j^l$ indicates whether the token is retained ($M_j^l > 0$) or masked out ($M_j^l = 0$). Specifically, 
for a generic token selection module $\mathcal{S}^l$, and for each token embedding \( \mathbf{h}_j^l \in \mathbf{H}^l \), the corresponding mask value is computed as:
\begin{equation}
    M_j^l = \text{ReLU}\left( l_g(\mathbf{h}_j^{l-1}) - l_t(\mathbf{b}^{l-1}) \right) \cdot M_j^{l-1}, 
    \label{eq:mask}
\end{equation}
for $j=1,\ldots,n-1$, where \( M_j^1 = 1 \) for all \( j \), indicating that all tokens are initially retained. For computational consistency, the mask values for the budget and class tokens are fixed to 1 at all layers. In (\ref{eq:mask}), by recursively multiplying the current mask with the one from the previous layer, any token that is discarded remains inactive for the remainder of the network, forcing consistency. The masking process can be applied either before or after each transformer block; in what follows, we assume the former. A visual overview of the proposed token selection mechanism is provided in \Cref{fig:zoom_gate}, highlighting how the model actively removes tokens at each stage. The corresponding pseudocode for the masking procedure is presented in \Cref{algo:code}. Finally, \Cref{fig:example_mask} illustrates an example of tokens being progressively discarded across consecutive blocks in the proposed adaptive token architecture, demonstrating its ability to retain only the semantically relevant information from the input image.

In the following section, we describe how the budget allocation process is trained to encourage the model to produce a number of tokens that closely matches the target budget $\alpha$.

%

\begin{algorithm}[t!]
\caption{Implementation of a Token selection module \( \mathcal{S} \)}
\label{algo:code}
\begin{algorithmic}[1]
\Require Set of token embeddings $\mathbf{H}$, budget token $\mathbf{b}$, previous mask $M$ (default: $M = 1$), threshold model $l_t(\cdot)$, gate model $l_g(\cdot)$ (we omit all layer indices for brevity).

\State $\text{threshold} \gets l_t(\mathbf{b})$   \Comment{Scalar threshold}
\State $\text{gates} \gets l_g(\mathbf{H})$   \Comment{One gate per token}
\State $m \gets \text{ReLU}(\text{gates} - \text{threshold})$
\State $m \gets m \times M$ \Comment{Apply previous mask recursively}
\State \Return $\mathbf{H} \cdot m,\; m$  \Comment{Mask the tokens}
\end{algorithmic}
\end{algorithm}



\subsection{Design of Token Selection Penalties and Training}

To enforce the computational budget constraints defined in Problem~\eqref{eq:formulation}, we introduce two complementary regularization terms. First, given a token selection module $\mathcal{S}^{l}$, with $l \le s$, we approximate the number of selected tokens by the average mask value, denoted by $\overline{M}^l$, and defined as:
\[
\overline{M}^l = \frac{1}{n+1} \sum_{j=1}^{n+1} M^l_j.
\]
where $M^l_j$ is calculated following \Cref{eq:mask}.
Then, we quantify the deviation from the target budget at block $l$ using the penalty function:
\begin{equation}
\label{eq:rc}
B(l) = \text{ReLU}\left( \left| \overline{M}^l - \alpha \right| - \epsilon \right), 
\end{equation}
where $\alpha$ is the target budget and $\epsilon$ is a margin parameter that allows tolerance in meeting the budget constraint. The ReLU function in (\ref{eq:rc}) ensures that penalties are incurred only when the deviation exceeds this margin. Then, from (\ref{eq:rc}), we impose the computational budget using two complementary losses. We define two regularization terms to guide the budget-aware training process:  
(i) the budget loss $B(s)$ in (\ref{eq:rc}), applied to the final adaptive block $s$, which enforces that the number of tokens transmitted over the channel matches the target budget; and  
(ii) a sparsity-inducing regularization $R$, which encourages early token elimination and is defined as:
\begin{equation}
\label{eq:ro}
R = \frac{1}{s} \sum_{l=2}^{s} B(l).
\end{equation}
The penalty term in (\ref{eq:ro}) promotes computational efficiency by encouraging the model to discard irrelevant tokens as early as possible in the network. Crucially, our recursive mask construction ensures that once a token is discarded, it remains inactive in all subsequent blocks, simplifying the optimization and enhancing training stability.


By incorporating the penalties in (\ref{eq:rc}) and (\ref{eq:ro}) and recasting the constrained problem (\ref{eq:formulation}) using the Lagrangian approach, the training objective of the proposed model reads as:
\begin{equation}\label{eq:unconstr_formulation}
\min_{\mathbf{w}}  \;\; \mathbb{E}_{(\mathbf{x}, y)}\,\mathcal{L}(f^{jscc}(\mathbf{x}, \alpha), y) + \lambda_s B(s) + \lambda_r R
\end{equation}
where $\lambda_s$ and $\lambda_r\geq 0$ are hyperparameters to balance task performance, budget adherence, and computational efficiency. Additionally, to allow for a correct training process, we initialize $l_g$ and $l_t$ so that, at the beginning of the training, their outputs are, respectively, around 1 and 0 for each input. This guarantees that at the beginning of the training no tokens are discarded, avoiding initialization biases. During training, we enable the model to handle varying computational budgets by sampling the budget parameter $\alpha$ uniformly from the interval $(0, 1]$ for each training instance. This approach effectively trains the model across the full spectrum of budget constraints, allowing it to adapt to any specified budget given by the user at inference time. At inference, we discretize the learned masks by removing tokens with zero-valued mask entries, thereby achieving actual computational savings.

\begin{figure*}[t!]
\centering
  \includegraphics[width=0.8\linewidth]{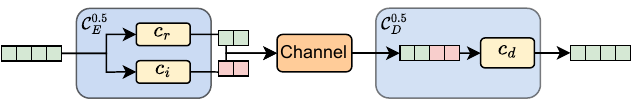}
  \caption{Overview of the proposed token-based transmission pipeline. The features of a generic token are processed by the endcoder $\mathcal{C}^r_E$, having a compression ratio of $r=0.5$, to produce the complex symbols, which are transmitted through a channel. The decoder $\mathcal{C}^r_D$ receives the symbols, concatenates imaginary and real parts, and recreates the features.} \label{fig:t-pipeline}
\end{figure*}

\subsection{Adaptive Token Compression}

Following iterative token selection, the last stage of the encoder \( E \) in (\ref{eq:encoder}) applies a token compression mechanism, denoted by \( \mathcal{C}_E: \mathbb{R}^{n_{\alpha} \times d} \rightarrow \mathbb{C}^{n_{\alpha} \times o} \), where \( n_{\alpha}=\mathcal{N}(\mathbf{x},\alpha) \) is the set of selected tokens, and \( o \leq d \). This module takes as input the subset of $n_{\alpha}$ selected tokens, produced by the encoder and transforms their features into a compact complex-valued representation suitable for transmission. Formally, the output of the encoder (after vectorization) yields the transmitted symbol vector \( \mathbf{s} \in \mathbb{C}^{q} \), with \( q = n_{\alpha} \cdot o \).  Specifically, the compression module \( \mathcal{C}_E \) consists of two components, \( c_R \) and \( c_I \), which generate the real part and the imaginary part of the complex-valued symbols, respectively. The resulting symbols are then normalized to satisfy the power constraint specified in (\ref{eq:Power_constraint}). At the receiver side, the decoder \( \mathcal{C}_D \) converts the received complex-valued signals into real-valued features. This is achieved by extracting the real and imaginary components of the input, concatenating them, and processing the result with a lightweight neural network \( c_D \). In our implementation, \( c_R \), \( c_I \), and \( c_D \) are all realized as three-layer feedforward networks, where each hidden layer is followed by a ReLU activation, except for the final output layer. The complete symbol autoencoding pipeline is illustrated in \Cref{fig:t-pipeline}.

As an additional degree of flexibility, we make the token feature compression adaptive via an online optimization strategy (cf. Section \ref{sec:alpha_tuning}), allowing the system to dynamically adjust the dimensionality of token representations based on their semantic importance and the available communication budget. To this end, as illustrated in Fig. \ref{fig:general-pipeline}, we employ a set of autoencoder pairs \( (\mathcal{C}_E^r, \mathcal{C}_D^r) \), each corresponding to a different compression ratio \( r = \frac{o_r}{e} \in (0, 1] \). Since \( r \) is tunable, the total number of complex symbols transmitted through the channel varies according to the selected autoencoder pair, becoming \( q_r = o_r \cdot n_{\alpha} \), where \( o_r \) is the output dimension associated with ratio \( r \), and \( n_{\alpha} \) is the number of selected tokens. The proposed DJSCC framework is therefore adaptable along two axes: the number of tokens selected for transmission, controlled by the parameter \( \alpha \), and the per-token dimensionality, controlled by the compression factor \( r \). In the next section, we exploit these degrees of freedom to jointly optimize system performance and communication efficiency.

\section{Dynamic Optimization of Token Selection and Compression}
\label{sec:alpha_tuning}
In the previous section, we presented the proposed token-based DJSCC mechanism, which provides design flexibility through two key parameters: the token budget $\alpha$ and the compression ratio $r$. In this section, we introduce a method for dynamically selecting the optimal pair $(\alpha, r)$ for DJSCC in time-varying scenarios, where channel conditions evolve over time due to factors such as fading, blockages, and user mobility. The method relies on a long-term problem formulation that is boiled down to a sequence of deterministic problems based on instantaneous observation of context parameters (e.g., channels) and opportunistically defined state variables.

We consider a time-slotted scenario where one image $\mathbf{x} \in \mathbb{R}^{p}$ is generated at each time instant $t$. Each image is first pre-processed locally using the token-based DJSCC, and the resulting intermediate representation is then transmitted wirelessly. Finally, the images are classified remotely using a decoder model. The wireless channel varies over time, and is assumed to be constant during a whole time slot (i.e., block fading). This setting requires a dynamic connect-compute DJSCC protocol that can adapt to changing conditions while maintaining stable performance. The adapted parameters are the compression ratio $r$ and the token budget $\alpha$, both of which are related to the ratio of radio resources allocated to the user. The objective is to maximize classification accuracy under bandwidth-related constraints.
To this aim, we define the following parameters: (i) the vector \( \boldsymbol{\gamma}(t) = \{r(t), \alpha(t)\} \), which is subject to optimization and consists of the compression ratio \( r(t) \in (0, 1] \) and the allocated budget \( \alpha(t) \in (0, 1] \); (ii) the compression factor \( \rho(\boldsymbol{\gamma}(t)) = \frac{q_{r,\alpha}(t)}{p} \), defined as the ratio of transmitted symbols \( q_{r,\alpha}(t) = o_r(t) \cdot n_\alpha(t) \) at time $t$ to the number of input symbols \( p \), where \( n_{\alpha}(t) \) denotes the number of selected tokens and \( o_{r}(t) \) the number of complex features per token, corresponding to the values of \( \alpha(t) \) and \( r(t) \), respectively; and (iii) \( \text{SNR}(t) \), the instantaneous signal-to-noise ratio at time slot \( t \). We further define: (iv) \( \Lambda(\boldsymbol{\gamma}(t),  \text{SNR}(t)) \), a proxy function representing the model’s accuracy as a function of the parameter configuration $\boldsymbol{\gamma}(t)\in\mathcal{C}$, and depending on the instantaneous channel condition via \( \text{SNR}(t) \); and (v) \( \rho_{\text{th}} \in (0, 1] \), a constraint on the compression factor, representing the maximum allowed average ratio of transmitted symbols relative to the total input size, typically determined by bandwidth limitations. 

The core objective of the algorithm is to dynamically select $\boldsymbol{\gamma}(t)$, i.e., the compression ratio \( r(t) \) and the budget \( \alpha(t) \), in order to maximize inference accuracy while satisfying a long-term constraint on the average number of transmitted symbols. Mathematically, this is equivalent to solving the long-term problem formulation:
\begin{equation}\label{prob_formulation}
\begin{split}
    \underset{\boldsymbol{\gamma}(t)\in \mathcal{C}}{\max} & \quad \lim_{T\to\infty}\frac{1}{T}\sum_{t=0}^{T-1}\mathbb{E}[\Lambda(\boldsymbol{\gamma}(t),  \text{SNR}(t))] \\
    &\text{subject to}  \;\;\lim_{T\to\infty}\frac{1}{T}\sum_{t=0}^{T-1}\mathbb{E}[\rho(\boldsymbol{\gamma}(t))]\leq \rho_{\text{th}}
\end{split}
\end{equation}
where the set \( \mathcal{C} \) denotes the discrete set of feasible values resulting from all possible combinations of \( r(t) \) and \( \alpha(t) \). The expectation is taken with respect to the channel realizations, whose statistics are assumed to be unknown a priori. To handle the long-term expectation constraint, we resort to Lyapunov stochastic optimization \cite{book}. We introduce a \textit{virtual queue} \( Z(t) \), which evolves according to the update rule:
\begin{equation}\label{eq:virtual_queue}
    Z(t+1) = \max\big(0,\, Z(t) + \mu(\rho(\boldsymbol{\gamma}(t)) - \rho_{\text{th}})\big),
\end{equation}
where \( Z(0) = 0 \), and \( \mu > 0 \) is a tunable step-size hyperparameter that controls the sensitivity of the queue dynamics.  Then, the long-term constraint can be equivalently reformulated in terms of the \textit{mean rate stability} of the virtual queue \( Z(t) \), i.e., \( \lim_{T \to \infty} \frac{\mathbb{E}\{Z(T)\}}{T} = 0 \). This stability condition ensures that the original long-term constraint is satisfied. By translating the constraint into a stability condition, the problem becomes more tractable. To proceed, we define a Lyapunov function \( \mathcal{L}(t) = \frac{1}{2}Z^2(t) \), and introduce the \textit{drift-plus-penalty} function:
\begin{equation}\label{DPP}
        \Delta_{p}(t \mid Z(t) ) = \mathcal{L}(t+1) - \mathcal{L}(t) - V \cdot \Lambda(\boldsymbol{\gamma}(t), \text{SNR}(t)),
\end{equation}
where \( V > 0 \) is a tunable parameter that balances the trade-off between optimizing performance and enforcing the constraint in \eqref{prob_formulation}. As shown in~\cite{book}, minimizing a suitable upper bound of \eqref{DPP} ensures the boundedness of the Lyapunov drift, which in turn guarantees the mean rate stability of \( Z(t) \), thereby satisfying the long-term constraint in (\ref{prob_formulation}). Based on this principle, we adopt a greedy slot-wise optimization strategy that minimizes the upper bound of \eqref{DPP} at each time slot. Following the derivations in~\cite{book} and~\cite{Merluzzi2021}, this approach leads to solving the following per-slot optimization problem, which depends only on the current channel observation and the virtual queue state:
\begin{align}\label{slot_problem}
    \underset{\boldsymbol{\gamma}(t)\in\mathcal{C}}{\min} \;\; & -V \cdot \Lambda(\boldsymbol{\gamma}(t), \text{SNR}(t)) + Z(t)\rho(\boldsymbol{\gamma}(t)).
\end{align}
Problem~\eqref{slot_problem} is significantly simpler than the original formulation in~\eqref{prob_formulation}, as it is deterministic at each time slot \( t \) and does not require prior statistical knowledge of contextual parameters (e.g., radio channel conditions). To solve this problem, we perform an exhaustive search over the set \( \mathcal{C} \), which typically has low cardinality, to identify the optimal pair of parameters \( \boldsymbol{\gamma}^*(t) = \{r^*(t), \alpha^*(t)\} \). Once the optimal solution is obtained, the virtual queue in (\ref{eq:virtual_queue}) is updated using the instantaneous compression ratio
$\rho(\boldsymbol{\gamma}^*(t)) = \frac{o_{r^*}(t) \cdot n_{\alpha^*}(t)}{p},$ where \( n_{\alpha^*}(t) \) denotes the number of selected tokens and \( o_{r^*}(t) \) the number of complex features per token, corresponding to the chosen values of \( \alpha^*(t) \) and \( r^*(t) \), respectively.  The pseudo-code that summarizes the proposed procedure is reported in \Cref{algo:opt}.

\begin{algorithm}[t!]
\caption{Dynamic Token Selection and Compression}
\label{algo:opt}
\begin{algorithmic}[1]
\Require  
A proxy function $\Lambda(\boldsymbol{\gamma}, {\rm SNR})$ for the accuracy, a constraint $\rho_{\text{th}}$, the hyperparameters $V$ and $\mu$, the feasible set $\mathcal{C}$ for the optimization parameters.
\State $Z(0) \gets 0$
\For{$t\geq 0$}
\State Observe ${\rm \text{SNR}}(t)$
\State $\gamma^*(t) \gets \underset{\boldsymbol{\gamma}(t)\in\mathcal{C}}{\arg\min} \;  -V \cdot \Lambda(\boldsymbol{\gamma}(t),  \text{SNR}(t)) + Z(t)\rho(\boldsymbol{\gamma}(t))$
\State $Z(t+1) \gets \max(0, Z(t) + \mu(\rho(\boldsymbol{\gamma}(t)) - \rho_{\text{th}}))$
\EndFor
\end{algorithmic}
\end{algorithm}

\noindent \textbf{Accuracy's proxy function and Hyperparameter selection.} To optimize the DJSCC system using \Cref{algo:opt}, we begin by constructing the proxy function \( \Lambda(\boldsymbol{\gamma},{\rm \text{SNR}}) \), which estimates the expected classification accuracy as a function of transmission parameters. Specifically, we systematically evaluate all combinations of compression ratio \( r \), computational budget \( \alpha \), and a discretized set of SNR values over the training dataset. For each configuration, we compute the average classification accuracy, thereby creating a mapping \( \Lambda \colon (\boldsymbol{\gamma}, \text{SNR}) \mapsto \text{Accuracy} \). This proxy enables efficient parameter selection during optimization by approximating model performance without requiring full inference. Following the construction of \( \Lambda \), we perform hyperparameter tuning by evaluating multiple combinations of the control parameters \( V \) and \( \mu \) via simulation. For each candidate pair, we execute the optimization procedure described in \Cref{algo:opt} over \( T = 10^5 \) simulation steps. We then compute the average compression ratio \( \Gamma(t) \) achieved over the final \( 10^3 \) steps of each run. The optimal hyperparameter configuration is selected as the one that maximizes the proxy accuracy \( \Lambda \), subject to the compression constraint. These optimized hyperparameters are then deployed in operational settings, where each input sample arrives at a specific time slot \( t \), and the channel conditions (SNR) may vary dynamically.

\section{Empirical evaluation}
\label{sec:results}
%
%

%
In this section, we evaluate our proposed DJSCC framework across multiple experimental settings\footnote{The code, used to run all the experiments, can be found in the \href{https://github.com/jaryP/AdaptiveSemanticTokenCommunication}{GitHub repository}.}. First, we describe our training methodology, detailing the model training, the integration of the communication pipeline, and the two distinct training scenarios: robust (with variable SNR) and noiseless. Next, we introduce our comprehensive set of baselines, including both neural network approaches and digital capacity-achieving methods. Finally, we present our experimental results and provide a detailed analysis of model performance across different channel conditions and compression settings.



\noindent{\textbf{Models pre-training.}}  Each model is pre-trained and fine-tuned on the Imagenette dataset \cite{imagenette}, which comprises $10,000$ images divided into $10$ classes. We evaluate three different architectures as backbones: a standard ViT \cite{vit}, a MobileNetV3 \cite{howard2019searching}, and our proposed Adaptive Semantic Token Selection approach.
All models are trained for 150 epochs with a batch size of 256, using the Adam optimizer \cite{kingma2014adam}. During training, we apply data augmentation techniques including RandAugment \cite{cubuk2020randaugment}, Color Jitter, and random horizontal flipping with probability $0.5$. For adaptive semantic token selection, we configure the first three blocks (i.e., $s=3$) to be adaptive while keeping the rest of the model static. We set the hyperparameters to $\lambda_s=2$ and $\lambda_r=1$ in (\ref{eq:unconstr_formulation}), and employ the budget sampling strategy detailed in Section \ref{sec:method}. This configuration provides an optimal trade-off between token reduction capability and classification accuracy.
\begin{figure*}[t]
    \centering
    \begin{subfigure}[t]{\linewidth}
        \includegraphics[width=0.3\linewidth]{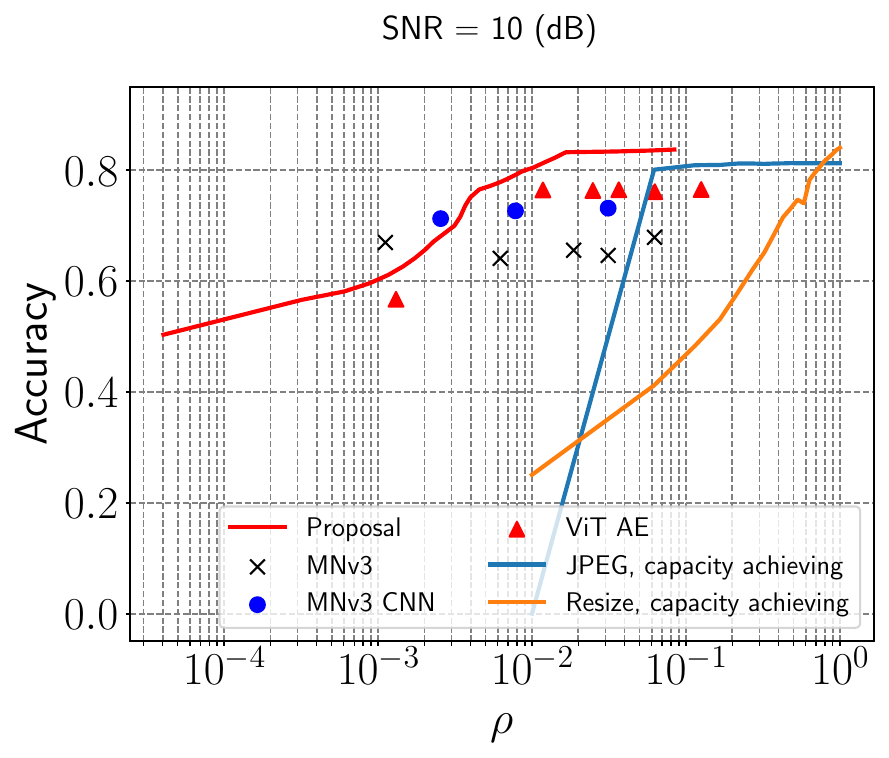}
    \hfill
        \includegraphics[width=0.3\linewidth]{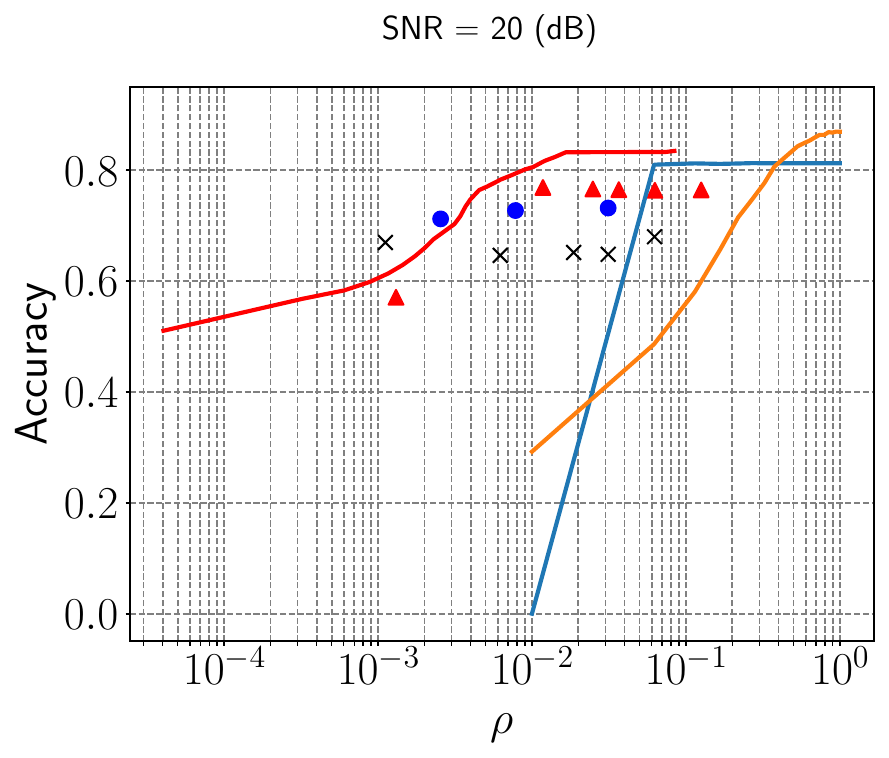}
    \hfill
        \includegraphics[width=0.3\linewidth]{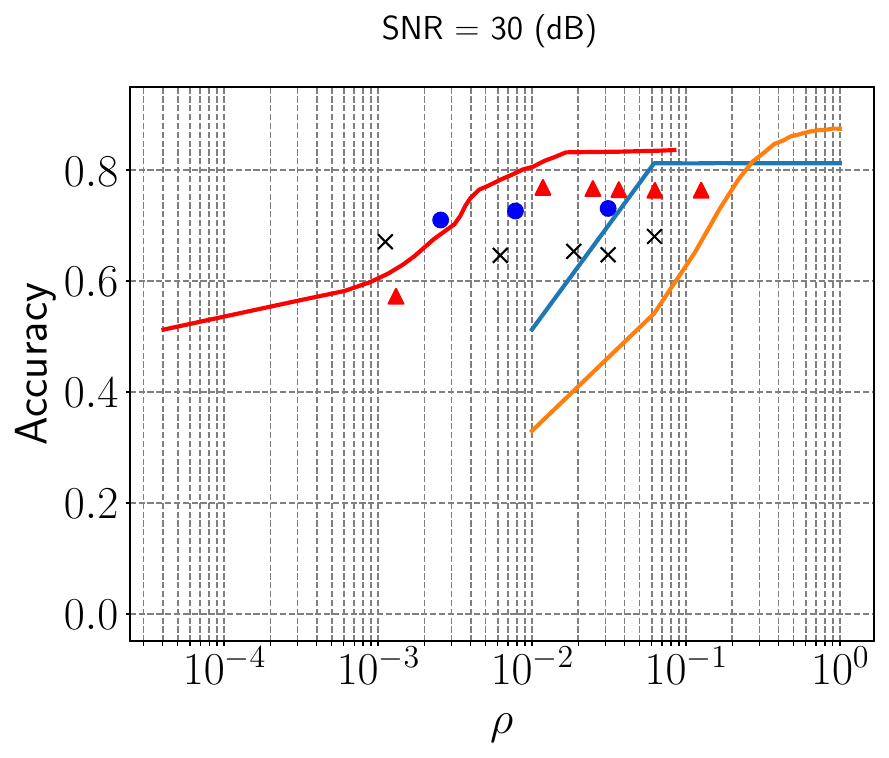}
    \caption{Accuracy curves when using robust DJSCC models.}
    \end{subfigure}
    \vfill
    \begin{subfigure}[t]{\linewidth}
        \includegraphics[width=0.3\linewidth]{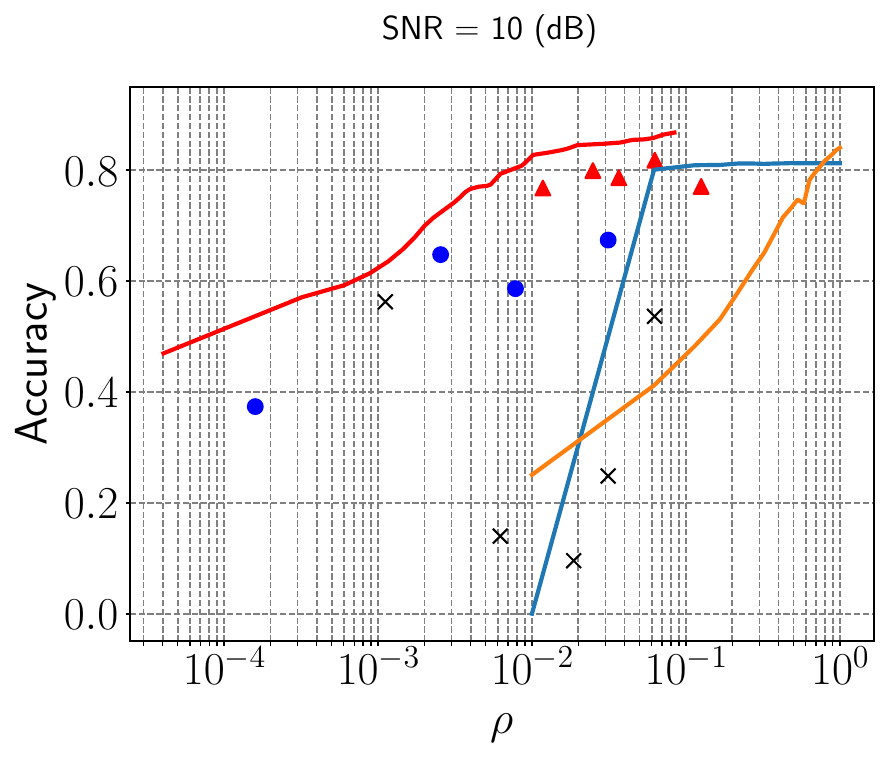}
    \hfill
        \includegraphics[width=0.3\linewidth]{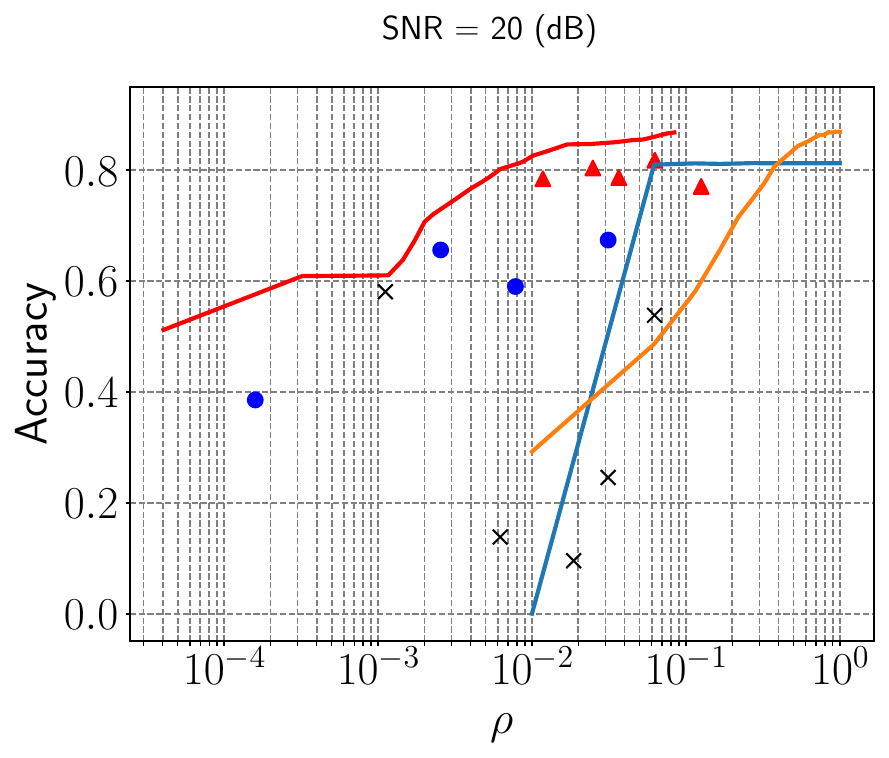}
    \hfill
        \includegraphics[width=0.3\linewidth]{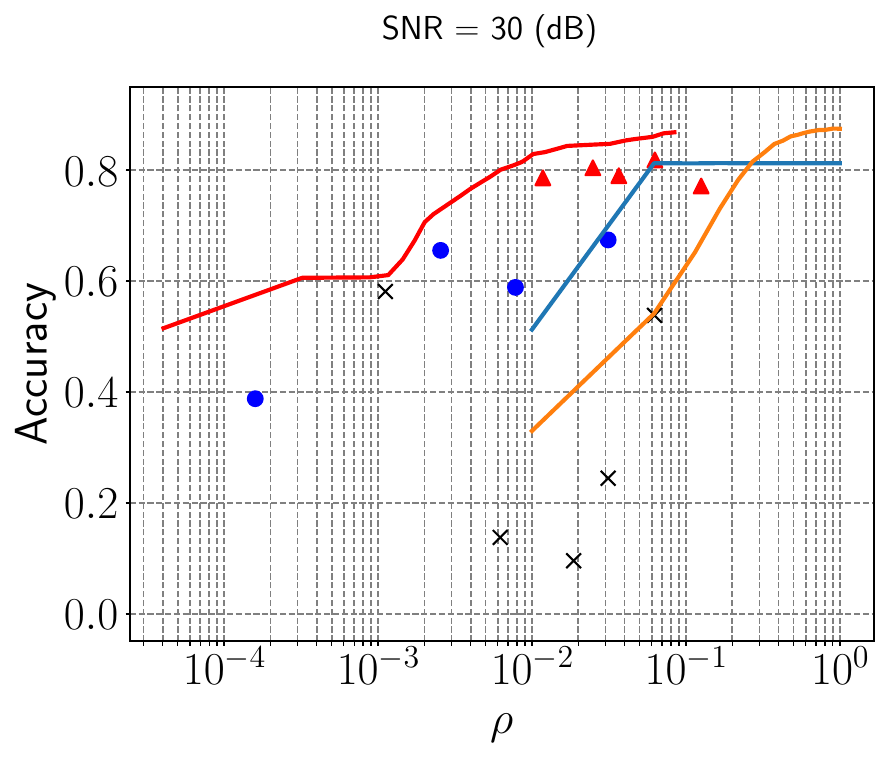}
    \caption{Accuracy curves when using noiseless DJSCC models.}
    \end{subfigure}
    \caption{Classification accuracy versus compression ratio \( \rho \) at different (high) SNR levels, comparing the proposed method with the baselines under two training settings: (a) robust training with noisy channels; and (b) training with ideal (noiseless) channels. The legend, shown in the upper-left plot, is shared across all subfigures.
    }
    \label{fig:high_snr}
\end{figure*}
\noindent\textbf{DJSCC set-up.} We convert each trained model into a DJSCC system by inserting a communication pipeline after the third block. Then, we fine-tune the complete DJSCC model 
%
exploring two distinct training scenarios. In the \textit{robust} training approach, for each training sample, we randomly sample an SNR value from a uniform distribution $\mathcal{U}[-20, 20]$ dB and apply the corresponding channel noise to the transmitted features. This approach acts as a regularization technique, ensuring that each sample is processed multiple times with different SNR values during training, making the model robust against channel perturbations. In contrast, in the \textit{noiseless} scenario, we do not apply any noise during training. However, we always apply noise during testing to evaluate model performance under variable channel conditions, regardless of the training approach used.
For a comprehensive evaluation, we test each DJSCC model with five different feature compression factors $r$: 0.005, 0.1, 0.15, 0.25, and 0.5. These varying compression levels allow us to analyze the trade-off between transmission efficiency and performance.

\noindent{\textbf{Baselines.}} We compare our proposed approach against two categories of baselines: neural network-based methods and digital capacity-achieving communication schemes. For our neural network comparisons, we implement three baseline models. Two are based on MobileNetV3 \cite{howard2019searching}: \textit{MNv3}, a standard MobileNetV3 where flattened features are converted to complex symbols by the DJSCC pipeline; and \textit{MNv3 CNN}, a variant using a convolutional DJSCC approach. For ViT, we implement \textit{ViT AE}, a static DJSCC ViT. For all such baselines, the model is described as in \Cref{eq:f_jscc}, and we use the same compression factors and splitting point ($s=3$) as described in the previous section.

Additionally, we consider two digital capacity-achieving communication baselines that compress and transmit the original image \( \mathbf{x} \in \mathbb{R}^{p} \) directly to the edge server, where classification is performed using a pretrained ViT model. According to Shannon's capacity formula, transmitting \( q \) symbols per image over a channel with a given SNR allows for a maximum of 
$
b_{\max} = q \log_2(1 + \mathrm{SNR})
$
bits to be reliably transmitted. If the image can be compressed to \( b_{\max} \) bits or fewer, we assume perfect reception at the receiver and pass the compressed image to the ViT for classification. Otherwise, the image is considered misclassified, reflecting the system’s inability to transmit it within the capacity constraint. The first baseline, \textit{Resize}, compresses the original \( H \cdot W \cdot 3 \) image into an \( L \cdot L \cdot 3 \) image, where each pixel channel is encoded using 8 bits. The total bit cost is therefore \( 24L^2 \). We then choose the largest integer \( L\geq 1 \) such that \( 24L^2 \leq b_{\max} \). If such $L$ cannot be found, the image is marked as misclassified.  The second baseline is \textit{JPEG}. In this case, we select the highest JPEG quality factor that results in a compressed image size (in bits) less than or equal to \( b_{\max} \). If no such quality factor can be found, the image is marked as misclassified.

\begin{figure*}[t]
    \centering
    \begin{subfigure}[t]{\linewidth}
        \includegraphics[width=0.3\linewidth]{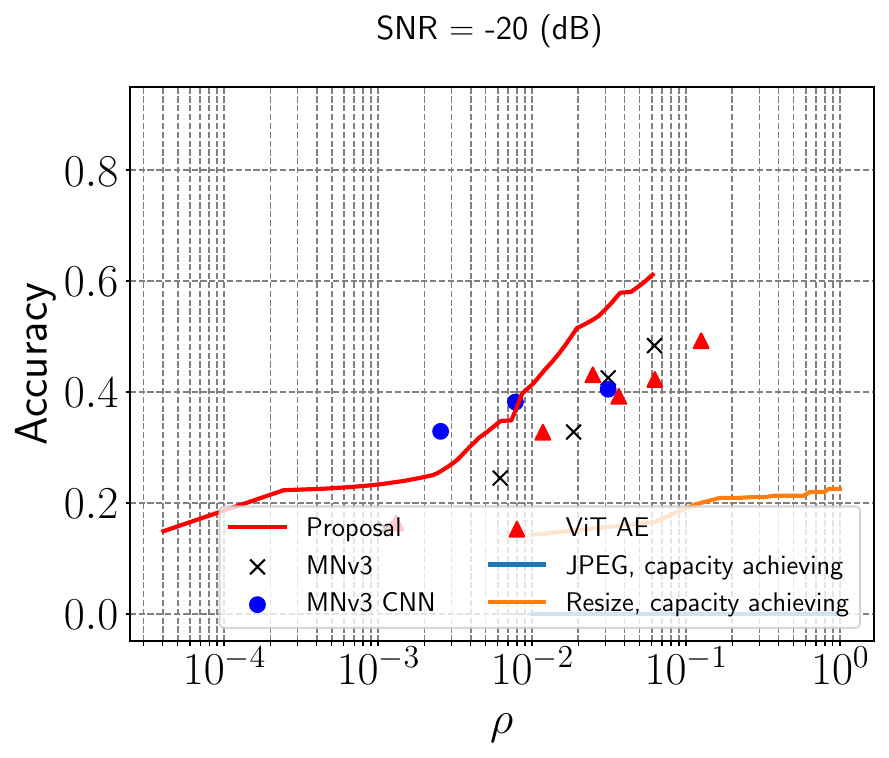}
    \hfill
        \includegraphics[width=0.3\linewidth]{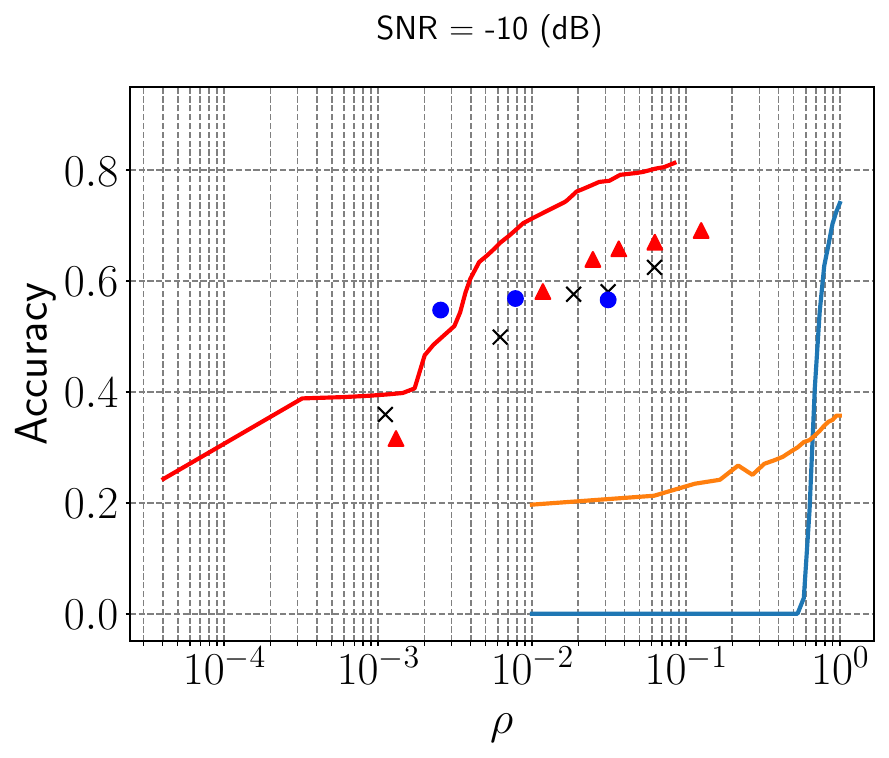}
    \hfill
        \includegraphics[width=0.3\linewidth]{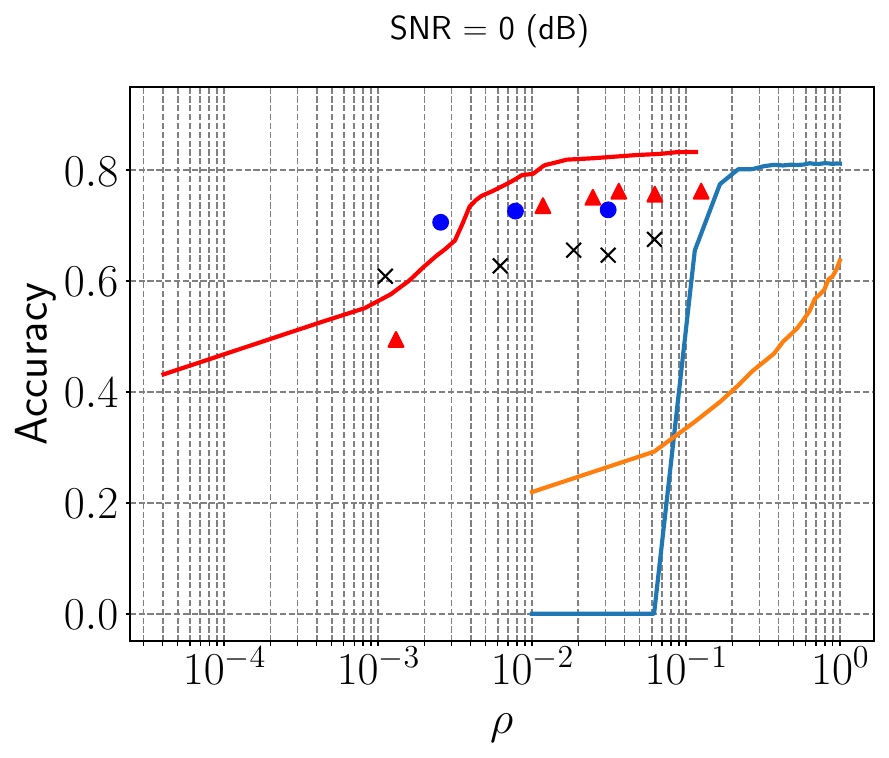}
    \caption{Accuracy curves when using robust DJSCC models.}
    \end{subfigure}
    \vfill
    \begin{subfigure}[t]{\linewidth}
        \includegraphics[width=0.3\linewidth]{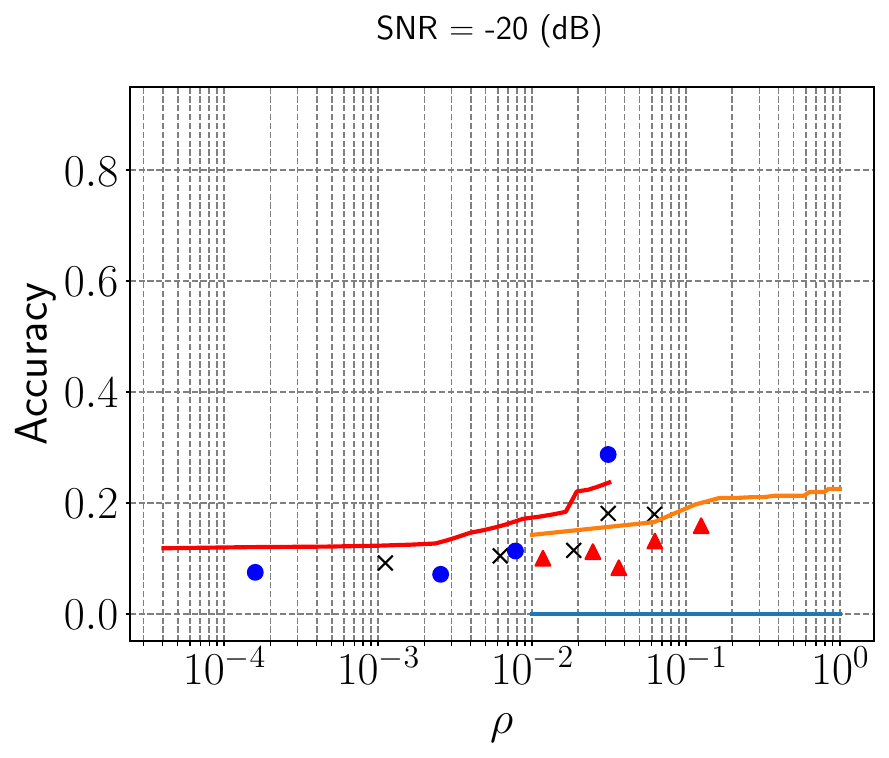}
    \hfill
        \includegraphics[width=0.3\linewidth]{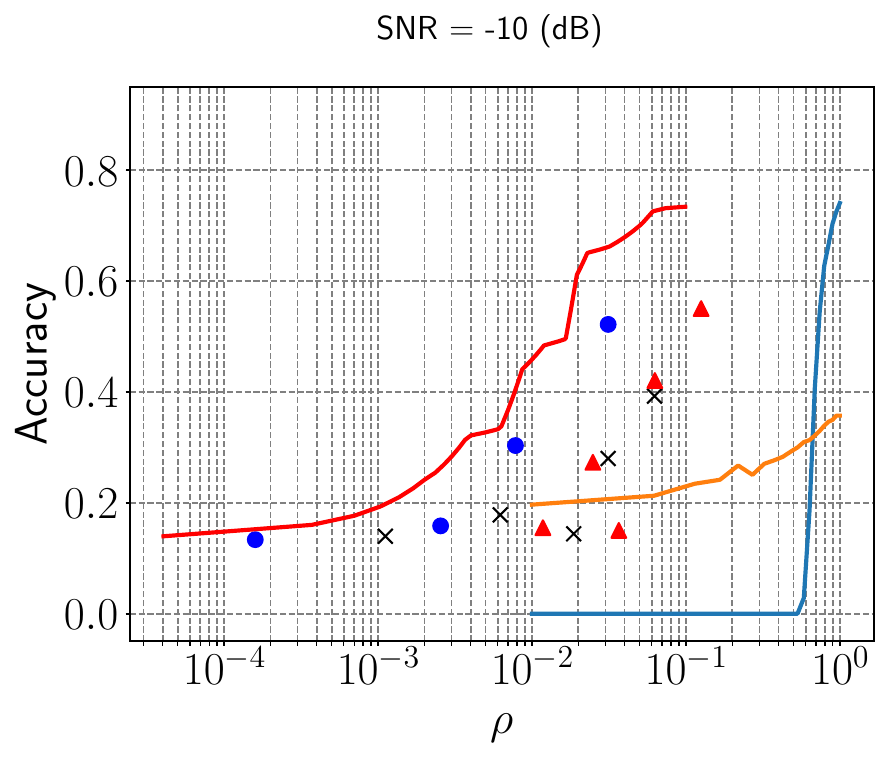}
    \hfill
        \includegraphics[width=0.3\linewidth]{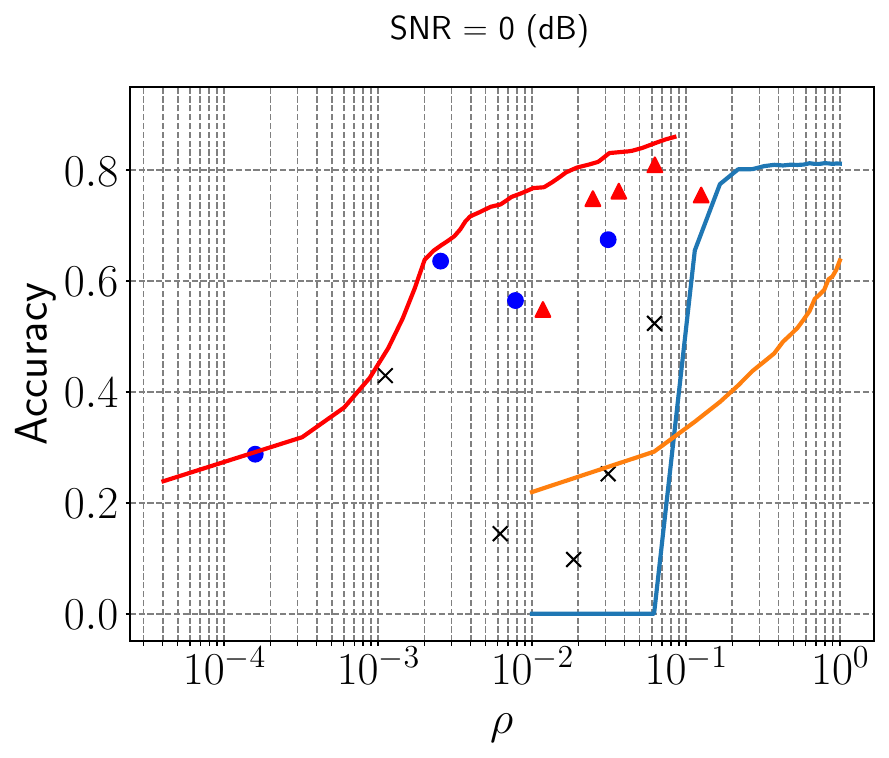}
    \caption{Accuracy curves when using noiseless DJSCC models.}
    \end{subfigure}
    \caption{Classification accuracy versus compression ratio \( \rho \) at different (low) SNR levels, comparing the proposed method with the baselines under two training settings: (a) robust training with noisy channels; and (b) training with ideal (noiseless) channels. The legend, shown in the upper-left plot, is shared across all subfigures.}
    \label{fig:low_snr}
\end{figure*}

\subsection{DJSCC performance with respect to noise and compression}

In this paragraph, we illustrate the performance of the proposed adaptive DJSCC model in terms of compression efficiency, considering both low and high SNR regimes, and compare it against the baseline methods. To this end, in \Cref{fig:high_snr} and \Cref{fig:low_snr}, we present the classification accuracy as a function of the compression ratio \( \rho \), comparing the proposed method with the baselines under high and low SNR conditions, respectively. Each figure considers two training scenarios: (a) training is performed over noisy channels with the corresponding SNR, and (b) training is done assuming ideal (noiseless) channels. In both cases, testing is conducted using the SNR specified in the respective subfigure. In these figures, the results for digital capacity-achieving communication schemes are shown as continuous lines, while the neural network baselines are depicted as discrete points, since each compression ratio requires training a separate DJSCC model, resulting in different \( \rho \) values. Our proposed method, on the other hand, enables a finer and more flexible selection of the final compression ratio \( \rho \), as a single model is trained to support multiple configurations by varying the input budget. This adaptability bridges the gap between the discrete compression ratios achievable by individual autoencoders. As a result, our method is represented by a single continuous curve, where only the best-performing configurations for each \( \rho \) value, in terms of classification accuracy, are shown. As we can notice from \Cref{fig:high_snr} and \Cref{fig:low_snr}, our method consistently outperforms all baselines, both digital and neural, achieving up to one or two orders of magnitude improvement in compression ratio compared to digital communication schemes for a given classification accuracy. As expected, the results obtained with robust training outperform those from training under ideal noiseless conditions, particularly in low and challenging SNR regimes. Conversely, at high SNR levels, the performance of both training strategies tends to yield similar results. Actually, from (\Cref{fig:high_snr}), we can notice how the models trained under ideal noiseless conditions slightly outperform their robustly trained counterparts. This is likely due to the SNR augmentation range used during training (set to [\(-20\), 20] dB), which may lead to performance flattening at the high end of the SNR spectrum. The adaptability of our proposal, combined with superior performance across different SNR regimes and budget constraints, demonstrates the practical advantages of our approach in real-world communication scenarios.

\subsection{Dynamic Optimization in Time-varying Channel Conditions}
\label{sec:tuning_alpha_res}
\begin{table*}[]
\centering
\resizebox{\textwidth}{!}{%
\begin{tabular}{|c|c||c|c|c|c|c|c|}
\cline{1-8}
\multirow{2}{*}{SNR (dB)} & \multirow{2}{*}{Method} & \multicolumn{6}{c|}{\multirow{2}{*}{$\rho_{\text{th}}$}} \\
 &  & \multicolumn{6}{c|}{} \\ \cline{3-8}
& & \multicolumn{1}{c|}{ 0.0025} & \multicolumn{1}{c|}{0.005} & \multicolumn{1}{c|}{ 0.01} & \multicolumn{1}{c|}{ 0.02} & \multicolumn{1}{c|}{ 0.05} & \multicolumn{1}{c|}{ 0.1}  \\ \hline \hline
%
%
\multicolumn{1}{|c|}{\multirow{3}{*}{$20$}} & ViT AE &  \multicolumn{1}{c|}{\result{\textbf{62.96}}{65.68}}&\multicolumn{1}{c|}{\result{68.05}{70.11}}&\multicolumn{1}{c|}{\result{\textbf{77.48}}{79.64}}&\multicolumn{1}{c|}{\result{\textbf{80.92}}{81.49}}&\multicolumn{1}{c|}{\result{\textbf{80.79}}{\textbf{83.40}}}&\multicolumn{1}{c|}{\result{80.97}{\textbf{84.09}}} \\ 
\cline{2-8} 
 \multicolumn{1}{|l|}{}  &  MNv3 CNN &  \multicolumn{1}{c|}{-}&\multicolumn{1}{c|}{\result{72.38}{65.38}}&\multicolumn{1}{c|}{\result{72.71}{65.91}}&\multicolumn{1}{c|}{\result{72.33}{66.65}}&\multicolumn{1}{c|}{\result{73.20}{67.36}}&\multicolumn{1}{c|}{\result{73.10}{67.46}} \\ \cline{2-8} 
 \multicolumn{1}{|l|}{}  &  Proposal &   \multicolumn{1}{c|}{\result{\textbf{63.41}}{\textbf{72.20}}}&\multicolumn{1}{c|}{\result{\textbf{71.14}}{\textbf{76.61}}}&\multicolumn{1}{c|}{\result{\textbf{78.93}}{\textbf{81.72}}}&\multicolumn{1}{c|}{\result{\textbf{81.15}}{\textbf{83.73}}}&\multicolumn{1}{c|}{\result{\textbf{81.07}}{\textbf{84.23}}}&\multicolumn{1}{c|}{\result{\textbf{85.15}}{\textbf{85.55}}}
 \\  \hline  \hline 
 \multicolumn{1}{|c|}{\multirow{3}{*}{$\mathcal{N}(20, 2.5)$}} & ViT AE &  \multicolumn{1}{c|}{\result{62.34}{65.84}}&\multicolumn{1}{c|}{\result{67.26}{70.67}}&\multicolumn{1}{c|}{\result{77.06}{\textbf{79.34}}}&\multicolumn{1}{c|}{\result{\textbf{80.92}}{\textbf{81.83}}}&\multicolumn{1}{c|}{\result{\textbf{80.54}}{82.55}}&\multicolumn{1}{c|}{\result{80.85}{83.25}}
 \\ 
\cline{2-8} 
 \multicolumn{1}{|l|}{}  &  MNv3 CNN &  \multicolumn{1}{c|}{-}&\multicolumn{1}{c|}{\result{72.10}{65.47}}&\multicolumn{1}{c|}{\result{72.75}{65.74}}&\multicolumn{1}{c|}{\result{72.65}{66.45}}&\multicolumn{1}{c|}{\result{73.15}{67.31}}&\multicolumn{1}{c|}{\result{73.16}{67.30}}
 \\ \cline{2-8} 
 \multicolumn{1}{|l|}{}  &  Proposal &   \multicolumn{1}{c|}{\result{\textbf{64.14}}{\textbf{71.90}}}&\multicolumn{1}{c|}{\result{\textbf{{74.73}}}{\textbf{76.66}}}&\multicolumn{1}{c|}{\result{\textbf{79.11}}{\textbf{79.97}}}&\multicolumn{1}{c|}{\result{\textbf{81.21}}{\textbf{81.48}}}&\multicolumn{1}{c|}{\result{\textbf{81.16}}{\textbf{84.39}}}&\multicolumn{1}{c|}{\result{\textbf{81.11}}{\textbf{85.52}}}
 \\ \hline  \hline 
\multicolumn{1}{|c|}{\multirow{3}{*}{$10$}} & ViT AE &  \multicolumn{1}{c|}{\result{\textbf{61.01}}{57.35}}&\multicolumn{1}{c|}{\result{\textbf{66.90}}{63.06}}&\multicolumn{1}{c|}{\result{77.20}{76.41}}&\multicolumn{1}{c|}{\result{\textbf{81.04}}{\textbf{83.41}}}&\multicolumn{1}{c|}{\result{\textbf{80.43}}{\textbf{83.63}}}&\multicolumn{1}{c|}{\result{\textbf{80.36}}{83.45}} \\ 
\cline{2-8} 
 \multicolumn{1}{|l|}{}  &  MNv3 CNN &  \multicolumn{1}{c|}{-}&\multicolumn{1}{c|}{\result{71.97}{65.35}}&\multicolumn{1}{c|}{\result{72.92}{65.45}}&\multicolumn{1}{c|}{\result{72.54}{66.14}}&\multicolumn{1}{c|}{\result{73.12}{67.41}}&\multicolumn{1}{c|}{\result{73.10}{67.31}}
 \\ \cline{2-8} 
 \multicolumn{1}{|l|}{}  &  Proposal &  \multicolumn{1}{c|}{\result{\textbf{62.88}}{\textbf{70.88}}}&\multicolumn{1}{c|}{\result{\textbf{66.23}}{\textbf{72.99}}}&\multicolumn{1}{c|}{\result{\textbf{79.31}}{\textbf{79.69}}}&\multicolumn{1}{c|}{\result{\textbf{81.10}}{\textbf{83.48}}}&\multicolumn{1}{c|}{\result{\textbf{81.07}}{\textbf{84.46}}}&\multicolumn{1}{c|}{\result{\textbf{80.82}}{\textbf{85.50}}}
  \\ \hline  \hline 
 \multicolumn{1}{|c|}{\multirow{3}{*}{$\mathcal{N}(10, 2.5)$}} & ViT AE &  \multicolumn{1}{c|}{\result{60.65}{57.49}}&\multicolumn{1}{c|}{\result{66.72}{64.53}}&\multicolumn{1}{c|}{\result{76.82}{77.19}}&\multicolumn{1}{c|}{\result{\textbf{80.22}}{81.02}}&\multicolumn{1}{c|}{\result{\textbf{80.69}}{\textbf{83.90}}}&\multicolumn{1}{c|}{\result{\textbf{80.34}}{83.26}} \\ 
\cline{2-8} 
 \multicolumn{1}{|l|}{}  &  MNv3 CNN &  \multicolumn{1}{c|}{-}&\multicolumn{1}{c|}{\result{71.83}{65.55}}&\multicolumn{1}{c|}{\result{72.70}{65.52}}&\multicolumn{1}{c|}{\result{72.97}{66.20}}&\multicolumn{1}{c|}{\result{73.09}{67.24}}&\multicolumn{1}{c|}{\result{73.00}{67.29}}
 \\ \cline{2-8} 
 \multicolumn{1}{|l|}{}  &  Proposal &  \multicolumn{1}{c|}{\result{\textbf{63.27}}{\textbf{69.90}}}&\multicolumn{1}{c|}{\result{\textbf{73.31}}{\textbf{74.51}}}&\multicolumn{1}{c|}{\result{\textbf{78.92}}{\textbf{79.16}}}&\multicolumn{1}{c|}{\result{\textbf{80.98}}{\textbf{83.04}}}&\multicolumn{1}{c|}{\result{\textbf{81.04}}{\textbf{84.36}}}&\multicolumn{1}{c|}{\result{\textbf{81.04}}{\textbf{85.03} }}
 \\ \hline  \hline 
%
 \multicolumn{1}{|c|}{\multirow{3}{*}{$\mathcal{N}(0, 2.5)$}} & ViT AE &  \multicolumn{1}{c|}{\result{\textbf{55.80}}{\textbf{63.80}}}&\multicolumn{1}{c|}{\result{61.07}{\textbf{72.79}}}&\multicolumn{1}{c|}{\result{\textbf{73.35}}{\textbf{79.04}}}&\multicolumn{1}{c|}{\result{\textbf{79.12}}{\textbf{80.72}}}&\multicolumn{1}{c|}{\result{80.02}{\textbf{81.45}}}&\multicolumn{1}{c|}{\result{\textbf{80.22}}{82.13}}  \\ 
\cline{2-8} 
 \multicolumn{1}{|l|}{}  &  MNv3 CNN &  \multicolumn{1}{c|}{-}&\multicolumn{1}{c|}{\result{71.41}{62.85}}&\multicolumn{1}{c|}{\result{72.63}{64.32}}&\multicolumn{1}{c|}{\result{72.79}{65.58}}&\multicolumn{1}{c|}{\result{72.95}{67.30}}&\multicolumn{1}{c|}{\result{72.93}{66.95}}
 \\ \cline{2-8} 
 \multicolumn{1}{|l|}{}  &  Proposal &   \multicolumn{1}{c|}{\result{\textbf{55.13}}{\textbf{64.80}}}&\multicolumn{1}{c|}{\result{\textbf{65.27}}{\textbf{73.00}}}&\multicolumn{1}{c|}{\result{\textbf{72.70}}{\textbf{78.73}}}&\multicolumn{1}{c|}{\result{\textbf{78.37}}{\textbf{79.88}}}&\multicolumn{1}{c|}{\result{\textbf{79.90}}{\textbf{82.04}}}&\multicolumn{1}{c|}{\result{\textbf{80.34}}{\textbf{83.81}}}

 \\ \hline 
\end{tabular}
}
\caption{For each combination of SNR (in dB, drawn from a distribution or a fixed point) and $\rho_{\text{th}}$, the accuracy calculated on the test is shown as \result{P}{R}, where P is the result of the minimization problem for noiseless communication pipelines, and R for the robust ones. Missing values correspond to failed minimization attempts (the $\rho$ value is higher than the threshold). The best values, and others with no more than 1\% accuracy difference from the best one, are highlighted in bold.}
\label{table:opt}
\end{table*} 
\begin{figure*}[t]
    \centering
    \begin{subfigure}[t]{\linewidth}
    \includegraphics[width=0.3\linewidth]{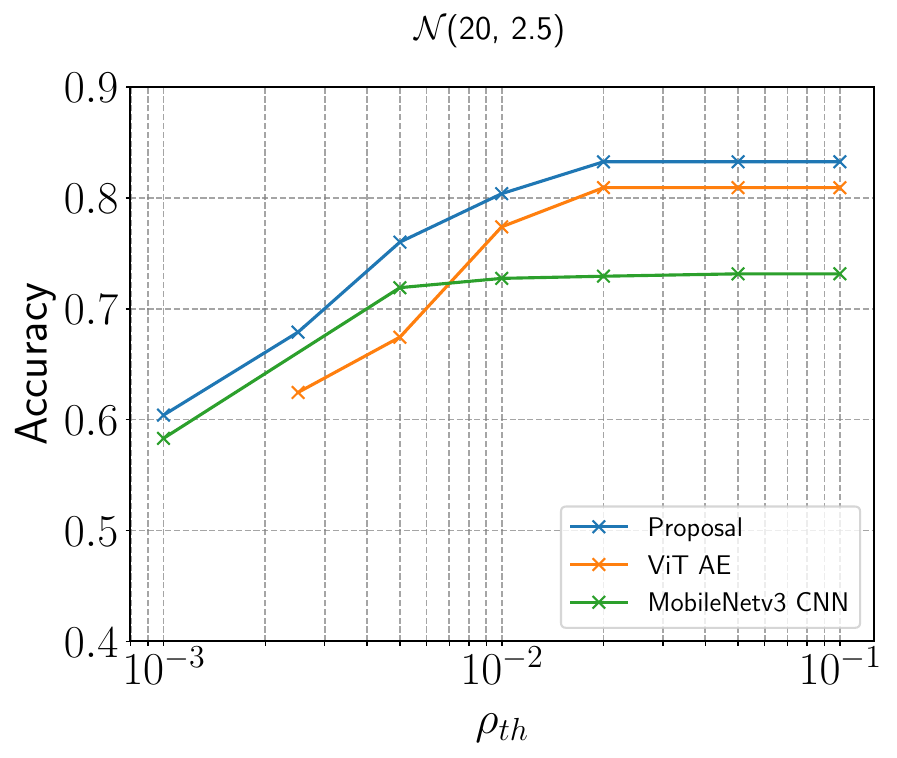}
    \hfill
    \includegraphics[width=0.3\linewidth]{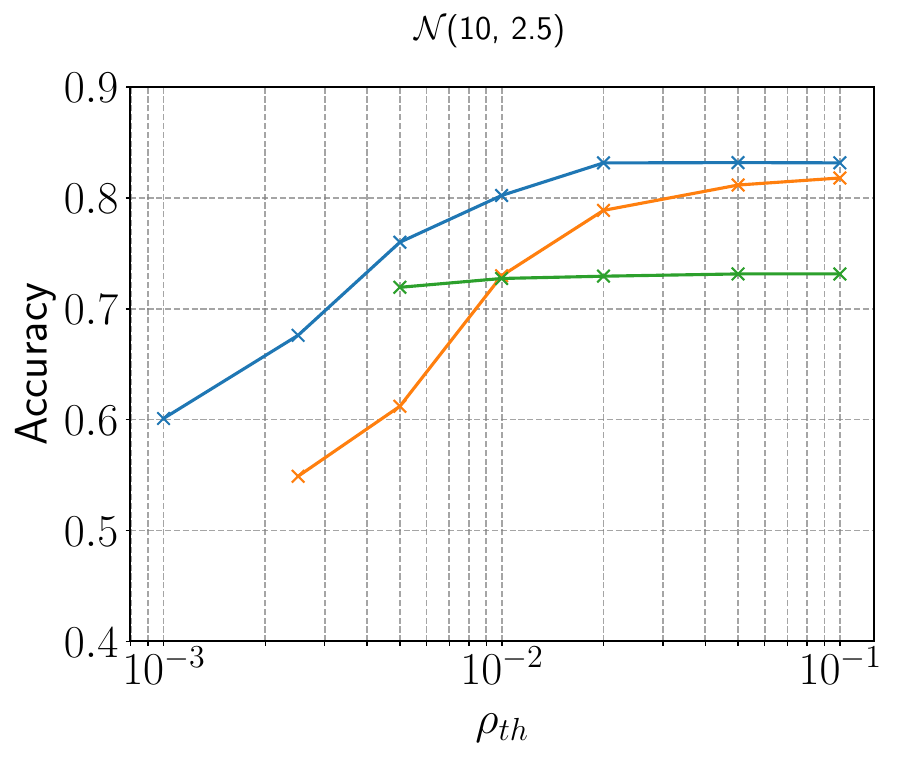}
    \hfill
    \includegraphics[width=0.3\linewidth]{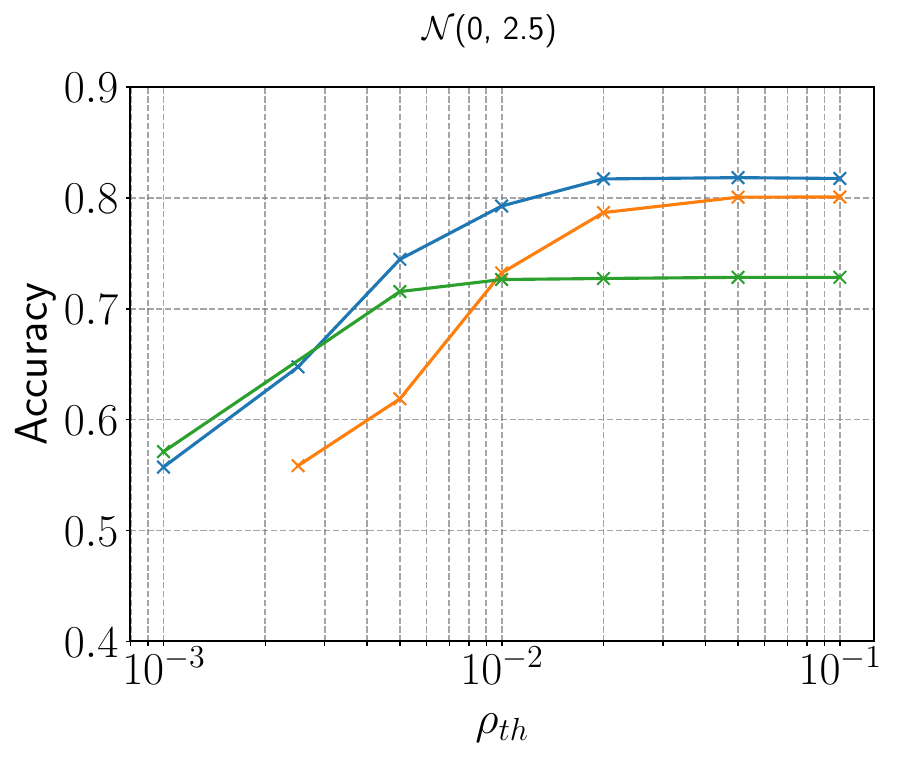}
    \caption{Accuracy curves when using robust JSCC models.}
    \end{subfigure}
    \vfill
    \begin{subfigure}[t]{\linewidth}
    \includegraphics[width=0.3\linewidth]{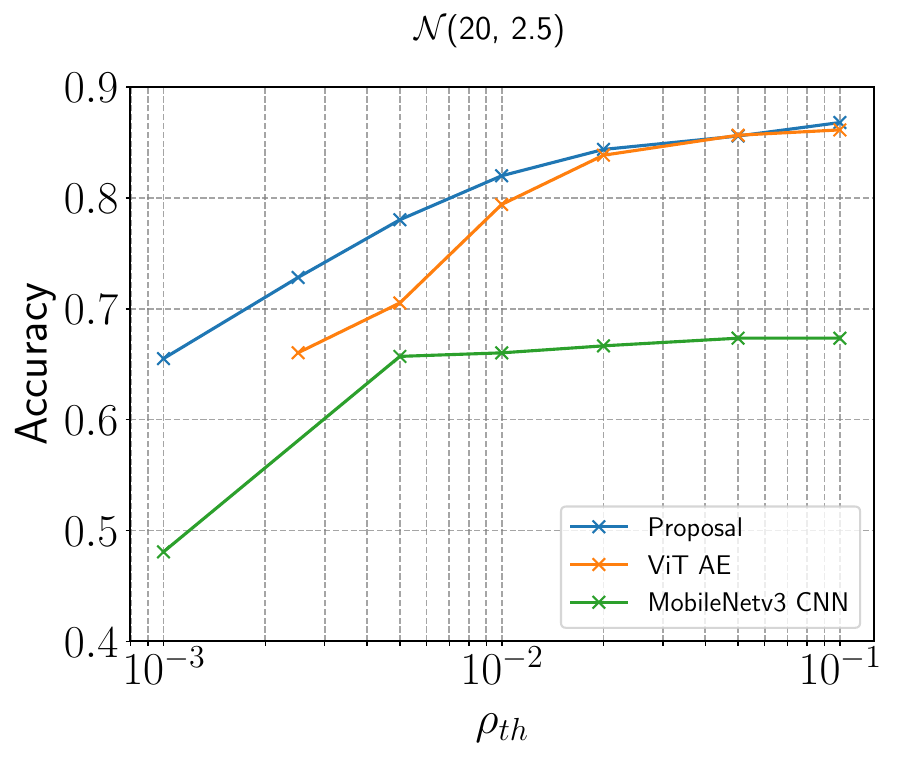}
    \hfill
    \includegraphics[width=0.3\linewidth]{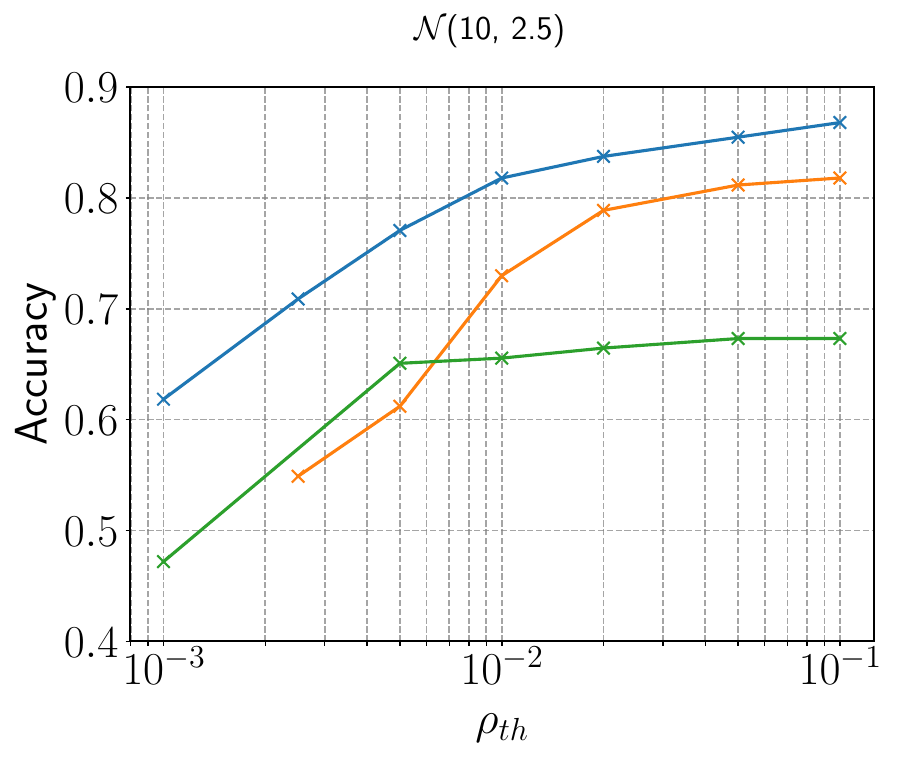}
    \hfill
    \includegraphics[width=0.3\linewidth]{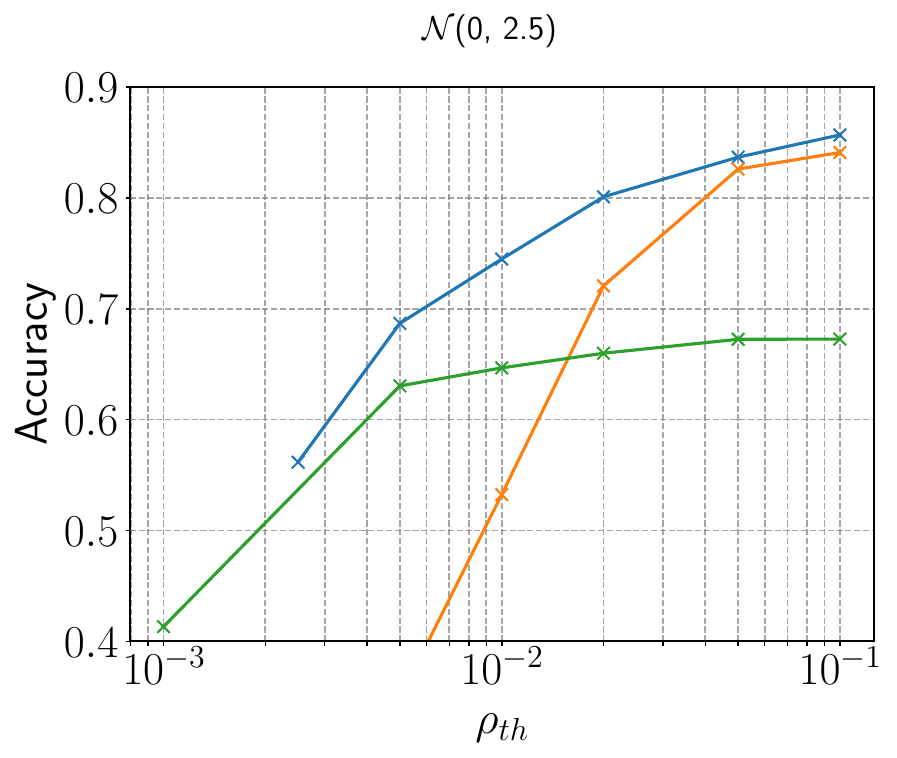}
    \caption{Accuracy curves when using noiseless DJSCC models.}
    \end{subfigure}
    \caption{The best results obtained for each minimization process are shown for different values of SNR (dB) distributions. On the top row, the results are obtained using robust JSCC models, while in the bottom one, the results are for noiseless ones. The legend is shared across all images.}
    \label{fig:opt}
\end{figure*}

In this section, we analyze the results obtained using the optimization approach proposed in Section~\ref{sec:alpha_tuning} for the dynamic adaptation of the selection and compression parameters \( \alpha(t) \) and \( r(t) \). We evaluated the proposed method across multiple scenarios. Specifically, for each threshold \( \rho_{\text{th}} \in \{0.005, 0.0025, 0.001, 0.02, 0.05, 0.01, 0.1\} \), we search for the optimal combination of the control parameters \( V \) and \( \mu \), with search spaces defined as \( V \in \{1, 10, 100, 1000, 10^4\} \) and \( \mu \in \{1, 10, 100\} \). To simulate different communication scenarios, we considered several SNR (dB) distributions: \( \mathcal{N}(10, 2.5) \), \( \mathcal{N}(20, 2.5) \), and \( \mathcal{N}(0, 2.5) \). In addition, we included two fixed SNR values: 10 dB and 20 dB. For scenarios involving statistical SNR distributions, we performed five independent optimization runs and report the average results. As baselines, we compare our proposed method against the previously introduced ViT AE and MobileNetV3 CNN models. For these DJSCC baselines, the only tunable parameter is the compression ratio \( r \).

To assess the performance of our approach, \Cref{fig:opt} presents representative optimization results under both robust and noiseless training regimes. On average, our method consistently outperforms the baselines across all scenarios. While ViT-AE achieves competitive performance in certain cases, our model maintains a clear advantage, particularly when robustness to channel noise is considered (\Cref{fig:opt}, top row). MobileNet, although it benefits from robust training, consistently underperforms compared to ViT-based approaches. Finally, \Cref{table:opt} reports the test accuracy across various SNR (dB) distributions and constraint values \( \rho_{\text{th}} \). Both robust and noiseless variants of the DJSCC models are evaluated. Overall, ViT-based models consistently achieve higher accuracy across all conditions. Notably, under stricter constraints (i.e., lower \( \rho_{\text{th}} \)), our proposed method outperforms all baselines, demonstrating its effectiveness in bandwidth-limited scenarios. As the constraint is relaxed, the performance gap between our model and ViT-AE narrows, reflecting the increased robustness to channel noise when more redundant features can be retained with lighter compression. In contrast, MobileNet models consistently underperform and often exhibit counterintuitive behavior, with accuracy degrading under robust training. This highlights their limited suitability for semantic transmission in noisy environments.

\subsection{Semantic Interpretability of Token Selection}

\begin{figure*}
\centering
\begin{tabular}{ccccc}
\subfloat{\includegraphics[width = 0.16\linewidth]{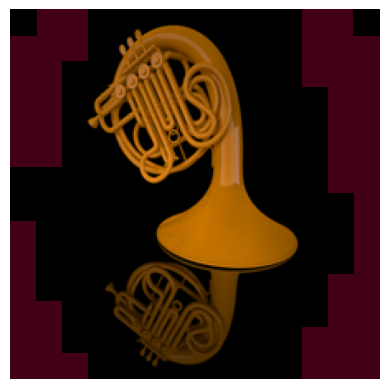}} &
\subfloat{\includegraphics[width = 0.16\linewidth]{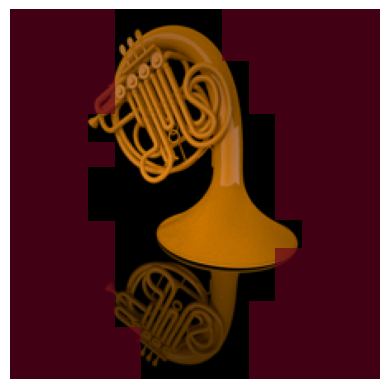}} &
\subfloat{\includegraphics[width = 0.16\linewidth]{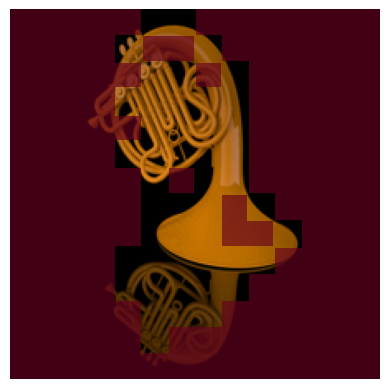}} &
\subfloat{\includegraphics[width = 0.16\linewidth]{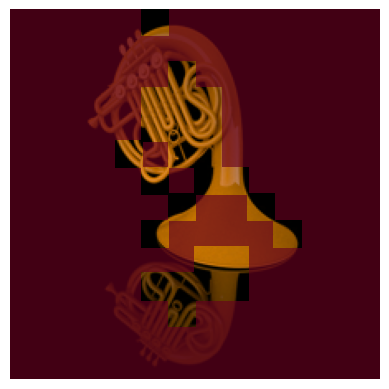}} &
\subfloat{\includegraphics[width = 0.16\linewidth]{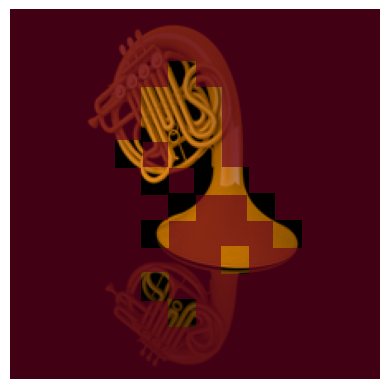}}\\
\subfloat{\includegraphics[width = 0.16\linewidth]{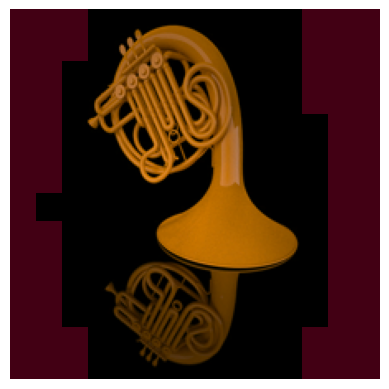}} &
\subfloat{\includegraphics[width = 0.16\linewidth]{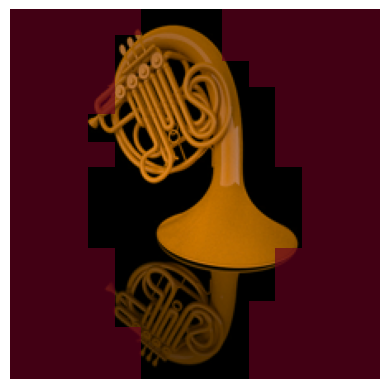}} &
\subfloat{\includegraphics[width = 0.16\linewidth]{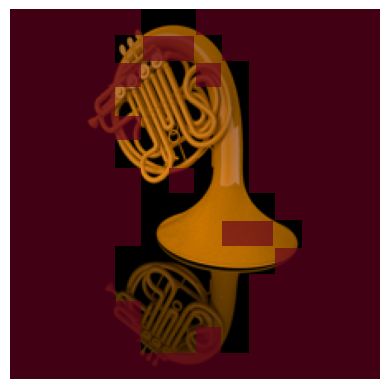}} &
\subfloat{\includegraphics[width = 0.16\linewidth]{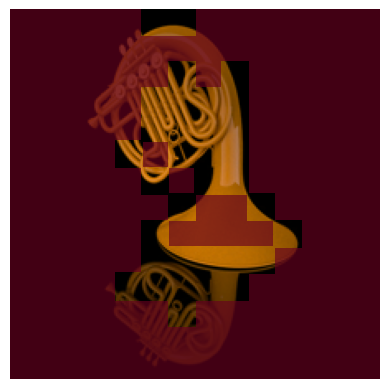}} &
\subfloat{\includegraphics[width = 0.16\linewidth]{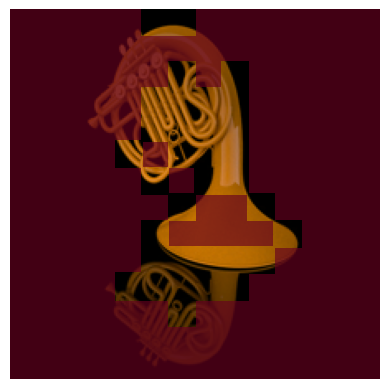}}\\
\subfloat{\includegraphics[width = 0.16\linewidth]{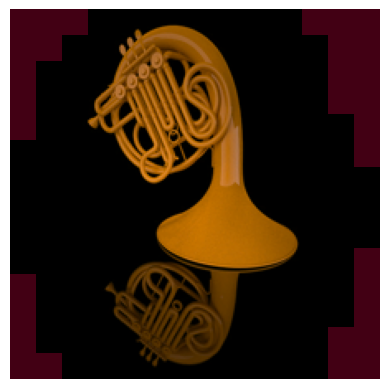}} &
\subfloat{\includegraphics[width = 0.16\linewidth]{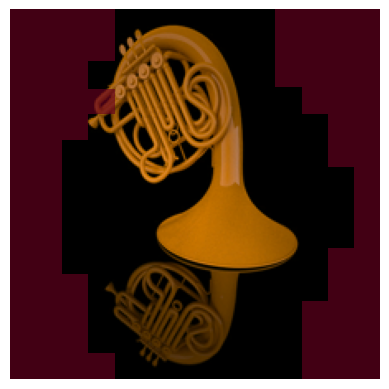}} &
\subfloat{\includegraphics[width = 0.16\linewidth]{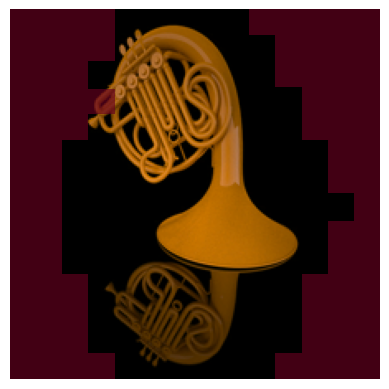}} &
\subfloat{\includegraphics[width = 0.16\linewidth]{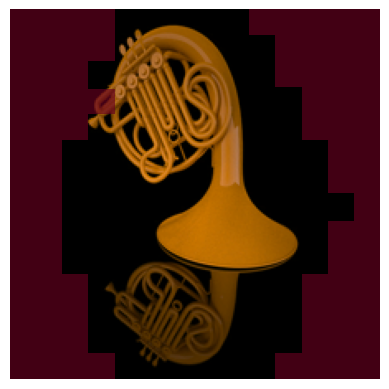}} &
\subfloat{\includegraphics[width = 0.16\linewidth]{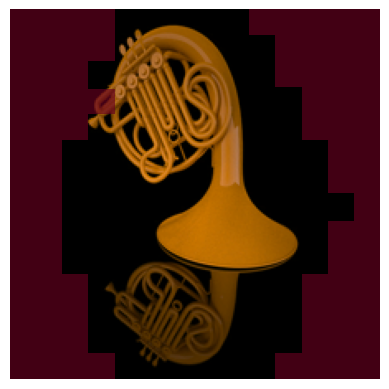}}\\
\end{tabular}
\caption{By decreasing the budget (from top to bottom), we increase the amount of discarded tokens (in red) at each block. The selected blocks, one per column, are 1, 2, 3, 5, and 12, while the selected budgets, one per row, are 0.1, 0.2, and 0.6.}
\label{fig:masks_2}
\end{figure*}

Our approach enables detailed visualization of token selection patterns across 
in 
\Cref{fig:masks_2}. These visualizations reveal a strong correlation between the retained tokens and semantically meaningful regions of the input, particularly under constrained computational budgets. In such settings, the model learns to prioritize tokens that convey task-relevant information, optimizing performance within limited resources. The token selection patterns highlight two notable phenomena. First, the model consistently favors tokens associated with semantically significant regions, demonstrating its ability to identify and preserve task-critical content. Second—and perhaps more intriguingly—certain tokens that may appear unimportant to human observers are consistently retained. This seemingly counterintuitive behavior can be attributed to the model's capacity to compress and redistribute information from discarded tokens into the retained ones, as similarly observed in \cite{darcet2023vision}.  This behavior indicates that the model develops sophisticated information preservation strategies that go beyond naive semantic filtering. These visualization capabilities not only offer insights into the model's internal decision-making process but also lay the groundwork for explainable-by-design AI-based communication systems.

%
%
%
\section{Conclusion and Future Work}
In this work, we introduced a novel transformer-based design for semantic edge inference, leveraging semantic token communication as a core mechanism. 

We propose a transformer-based semantic token selection mechanism that dynamically identifies and transmits task-relevant input components, such as image patches. The architecture compresses selected tokens into complex-valued representations with minimal reconstruction loss, maintains robustness across diverse channel conditions using a single model, and remains compatible with standard real-valued neural networks for seamless integration with pretrained models. By reducing the number of tokens processed and transmitted, it lowers computational and communication overhead, making it suitable for resource constrained edge environments. Additionally, a stochastic optimization-based resource allocation algorithm adapts system parameters in real time to balance inference accuracy and communication cost under varying conditions, enabling efficient and adaptive semantic transmission. Supported by a robust and flexible DJSCC pipeline, our approach enables adaptive handling of communication constraints while preserving high accuracy. Our experimental results demonstrate that the proposed model consistently outperforms existing baselines, delivering superior accuracy across diverse operational constraints while maintaining robustness to channel impairments. A significant feature of our approach is its contribution to interpretable AI-native communication systems through the inherent transparency of the semantic token selection process. The ability to visualize and understand token selection patterns provides valuable insights into the system's decision-making process. 

As future developments, we aim to extend this work across multiple data modalities, as we hypothesize that certain tasks may benefit from even more compact representations using fewer tokens. Additionally, we plan to explore the integration of trainable empty tokens, which could serve as a form of memory within the communication process—facilitating more sophisticated information storage and retrieval post-transmission. We also intend to investigate multi-user communication scenarios, where efficient resource allocation—such as bandwidth, power, and token budgets—becomes critical. In such settings, the semantic communication framework must balance the needs and priorities of multiple users while preserving overall system performance. These enhancements have the potential to further improve the flexibility, scalability, and expressiveness of semantic communication systems.

\section*{Acknowledgments}
This work has been supported by the SNS JU project 6G-GOALS under the EU’s Horizon program Grant Agreement No 101139232, by Sapienza grant RG123188B3EF6A80 (CENTS), and by European Union under the Italian National Recovery and Resilience Plan of NextGenerationEU, partnership on Telecommunications of the Future (PE00000001 - program RESTART).  We also acknowledge ISCRA for awarding this project access to the LEONARDO supercomputer, owned by the EuroHPC Joint Undertaking, hosted by CINECA (Italy).

\bibliographystyle{IEEEtran}
\bibliography{biblio}

\end{document}